\pdfoutput=1
\documentclass[11pt]{article}

\usepackage[]{emnlp2023}

\usepackage{times}
\usepackage{latexsym}

\usepackage[T1]{fontenc}
\usepackage[utf8]{inputenc}

\usepackage{microtype}
\usepackage{inconsolata}
\usepackage{booktabs}
\usepackage{xspace}
\usepackage{graphicx}
\usepackage{multirow}
\usepackage{makecell}
\usepackage{colortbl}
\usepackage[normalem]{ulem}
\usepackage{algorithm}
\usepackage[noend]{algpseudocode}
\usepackage{amsmath}
\usepackage{amssymb}
\usepackage[]{xcolor}

\definecolor{cyan_c}{RGB}{222, 255, 255}

\definecolor{violet-5}{RGB}{132, 94, 247}
\newcommand{\dataset}{\textsc{$\delta$-Rules-of-Thumb}\xspace}
\newcommand{\datasetshort}{\textsc{$\delta$-RoT}\xspace}

\newcommand{\studentbase}{\textbf{Distill}$_{\text{base}}$\xspace}
\newcommand{\studentone}{\textbf{SelfDistill$_1$}\xspace}
\newcommand{\studenttwo}{\textbf{SelfDistill$_2$}\xspace}
\newcommand{\studentfinal}{\textbf{Distill}$_{\text{final}}$\xspace}

\newcommand{\socialchem}{\textsc{Social-Chem-101}\xspace}

\newcommand{\eg}{\textit{e.g.},\xspace}
\newcommand{\ie}{\textit{i.e.},\xspace}

\title{What Makes it Ok to Set a Fire? Iterative Self-distillation of Contexts and Rationales for Disambiguating Defeasible Social and Moral Situations}

\newcommand{\aspace}{\hspace{1em}}
\newcommand{\uw}{$^{\heartsuit}$}
\newcommand{\aitwo}{$^{\spadesuit}$}
\newcommand{\equalcontribute}{$^{*}$}

\author{%
    \textbf{Kavel Rao}\uw \equalcontribute \aspace 
    \textbf{Liwei Jiang}\uw \aitwo \thanks{~\, Equal contribution.} \aspace 
    Valentina Pyatkin\aitwo \aspace
    Yuling Gu\aitwo \aspace \\
    \textbf{Niket Tandon}\aitwo \aspace
    \textbf{Nouha Dziri}\aitwo \aspace
    \textbf{Faeze Brahman}\aitwo \aspace
    \textbf{Yejin Choi}\uw \aitwo \aspace \\
    \uw{}Paul G. Allen School of Computer Science \& Engineering, University of Washington \\
    \aitwo{}Allen Institute for Artificial Intelligence \\
    \texttt{\{kavelrao,lwjiang\}@cs.washington.edu}
}

\begin{document}

\maketitle

\begin{abstract}

% \nouha{The abstract is a bit long for a short paper. }
% Moral or ethical judgments are often highly \textit{context} dependent. For instance, ``knowing where someone lives'' is a morally neutral action,
% % \kavel{morally neutral action? or morally ambiguous action?}, 
% while providing the additional context ``to provide assistance to a person in need'' makes the action \textit{more} morally acceptable. However, providing the context ``to spy on them'' makes the action \textit{less} justified. 
% \nouha{This example is a candidate for pruning, examples can go to intro}

Moral or ethical judgments rely heavily on the specific contexts in which they occur.
Understanding varying shades of \textit{defeasible contextualizations} (\ie additional information that strengthens or attenuates the moral acceptability of an action) is critical to accurately represent the subtlety and intricacy of grounded human moral judgment in real-life scenarios.

We introduce \textit{defeasible moral reasoning}: 
% a task that involves providing grounded contexts that make a given action more or less morally acceptable, along with commonsense rationales that justify the reasoning.
a task to provide grounded contexts that make an action more or less morally acceptable, along with commonsense rationales that justify the reasoning.
% \faeze{and rationalization?} \liwei{I'm a bit hesitant about highlighting rationalization as we don't do much about the rationale part other than distill from the teacher model. Kind of leaning towards fold that into the concept of defeasible moral reasoning} \faeze{fair}
% task: give a situation to provide contexts that make a situation more or less morally acceptable and the rationales on why so. \faeze{provide additional context that makes a given situation more or less morally acceptable and the rationales on why so.} \faeze{alternative: We introduce \textit{defeasible moral reasoning and rationalization}. The task involves providing additional context that makes a given situation more or less morally acceptable, along with the corresponding rationales that justify the reasoning.} \kavel{whether including rationales or not, i like the wording in Faeze's alternative} 
% To elicit a high-quality dataset for this task
To elicit high-quality task data, we take an iterative self-distillation approach that starts from a small amount of unstructured seed knowledge from GPT-3 and then alternates between (1) self-distillation from student models; (2) targeted filtering with 
% human judgment approximation critic scores \faeze{do we mean approximated human judgment critic score? this phrase is quite unclear.}
a critic model trained by human judgment (to boost validity) and NLI (to boost diversity); (3) self-imitation learning (to amplify the desired data quality). This process yields a student model that produces defeasible contexts with improved \textit{validity}, \textit{diversity}, and \textit{defeasibility}. From this model we distill a high-quality dataset, \dataset (\datasetshort), of 1.2M entries of contextualizations and rationales for 115K defeasible moral actions rated highly by human annotators 85.9\% to 99.8\% of the time.\footnote{Dataset is publically available at \url{https://huggingface.co/datasets/kavelrao/d-Rules-of-Thumb}} Using \datasetshort we obtain a final student model that wins over all intermediate student models by a notable margin.

\end{abstract}

\section{Introduction}

\begin{figure}[t!]
    \includegraphics[width=\linewidth]{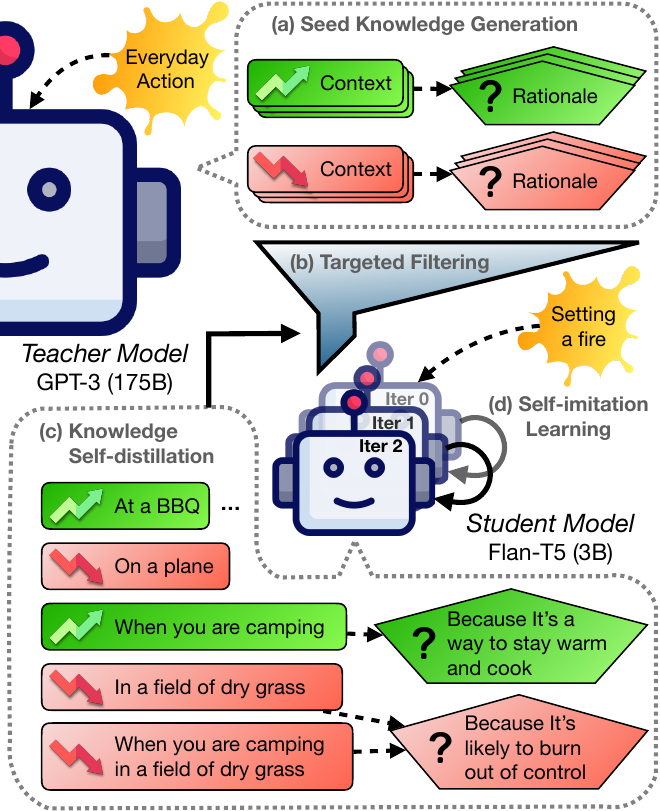}
    \caption{An illustration of the iterative self-distillation pipeline on eliciting \textit{defeasible moral reasoning} \textit{contexualizations} and \textit{rationales}. 
    % Starting from a small set of seed knowledge (a) elicited from GPT-3 and (b) filtered for targeted property, we amplify learning signals by iteratively training smaller student models by repeating (c) knowledge self-distillation; (b) targeted filtering; (d) and self-imitation learning. 
    For the event ``setting a fire,'' different contextualizations of the action can bend its moral acceptability \textit{up} (\eg ``at a BBQ'') or \textit{down} (\eg ``to get revenge''). Capturing the nuances of how additional contexts interplay with base actions is critical for grasping the flexible defeasibility of moral judgments.}
    \label{fig:open-fig}
    \vspace{-6mm}
\end{figure}

% ``when you're camping,'' ``in a chemistry procedure'') 
% ``on a plane,'' ``in a field of dry grass''). 

 % \nouha{the process of self-distillation pipeline is not clear for people who are not familiar with the concept. I recommend adding numbers to clarify which step comes first. }

Moral or social judgments play a vital role in decision-making, influencing how we perceive actions and behaviors daily. However, these judgments are far from fixed; instead, they are highly \textit{context-dependent}. Contexts surrounding a core action can significantly \textit{strengthen} or \textit{weaken} its moral acceptability. For instance, the act of ``knowing where someone lives'' may carry no inherent moral weight. But when supplemented by the context that ``the purpose is to provide assistance to a person in need,'' the action becomes more morally justified. Conversely, if the context shifts to ``for the purpose of surveillance or spying,'' the same action loses its moral grounding.
This phenomenon of flexibly bending moral rules in instantiations of scenarios is widely recognized in assorted cognitive science studies
\cite{kwon2022flexibility,levine2020logic,awad2022acceptable,Levine2018TheCM}.

The inherent context dependence of moral judgments underscores the importance of understanding the complex interplay between actions and their grounded contexts in real-life scenarios.
% The importance of understanding the complex interplay between actions and their grounded contexts in real-life scenarios is underscored by the inherent context dependence of moral judgments.
% Understanding what contexts changes moral implications of an action in what way and why enables us to make informed moral judgments geared toward situational nuances.
Delving into how different contexts bend the moral acceptability of an action, along with the reasons behind these shifts, enables us to make informed moral judgments geared toward situational nuances.
% empowers us to make well-informed moral judgments that take into account the nuances of each situation.
% To this end, we introduce the concept of \textit{defeasible moral reasoning}. The goal of the task is to provide grounded \textit{contextualizations} (or contexts) that alter the moral acceptability of action, accompanied by commonsense \textit{rationales} that justify the reasoning. 

Previous works about contextualized moral judgment pose several challenges. First, they focus primarily on atomic contexts with limited situational complexity. For instance, \citet{ziems2023normbank} rigidly prescribe grounded contexts to fall under concepts such as settings, roles, and behaviors, narrowing and fragmenting the scope of contextualization. \citet{pyatkin2023clarifydelphi} propose a stepwise clarification question generation system to elicit elementary contexts of moral actions. 
% Another limitation lies in the emphasis on the defeasibility of assumed moral and social judgments. 
Another limitation lies in the emphasis on the defeasibility of assumed moral and social judgments (\eg ``it's wrong to yell to your friend''), 
% They utilize predefined norms such as ``it's wrong to make loud noise late at night'' 
rather than the natural defeasibility of moral scenarios themselves \cite{rudinger-etal-2020-thinking}.
Finally, existing works lack the rationales to explain why a particular context renders a situation more or less morally acceptable. We address all the above limitations in this work.

We introduce \textit{defeasible moral reasoning}, a task to provide grounded \textit{contextualizations} (or contexts) that alter the moral acceptability of action, accompanied by commonsense \textit{rationales} that justify the reasoning. We aim to explicitly state underspecified contexts of actions, providing nuance and interpretability to moral judgments.
To substantiate this task, we introduce \dataset (\datasetshort), a high-quality dataset of 1.2M entries of combined contextualizations and rationales for 115K defeasible moral actions. \datasetshort is created through an iterative self-distillation approach. Starting with a small amount of unstructured seed knowledge from GPT-3, we alternate between (1) \textbf{self-distillation from student models} to move away from the reliance on expensive API calls for GPT-3; (2) \textbf{targeted filtering} using a critic model trained with human judgment to enhance generation validity and natural language inference (NLI) to enhance diversity; and (3) \textbf{self-imitation learning} to magnify the desired model properties. 

The iterative self-distillation process yields a student model that generates defeasible contexts with enhanced \textit{validity}, \textit{diversity}, and \textit{defeasibility}.
From the best-performing student model, we distill the final dataset, \datasetshort. These contextualizations and rationales have been rated highly by human annotators on both validity (85.9\%) and language quality (99.8\%), ensuring their reliability in capturing the complexities of moral judgment within various contexts. Using \datasetshort, we further train a final downstream student model that prevailed over all intermediate student models.

In sum, in this work, we introduce the defeasible moral reasoning task that involves contexts and rationales for making defeasible moral judgments. We present the iterative self-distillation methodology to gather high-quality and diverse training data for student models (\S\ref{sec:iterative}) along with careful ablations of each component involved in the pipeline (\S\ref{sec:results}). We distill a sizeable, human-verified dataset under the defeasible moral reasoning task formulation (\S\ref{sec:data}) and a high-performing downstream task model. We will release all student models and \dataset, along with a subset of human-annotated gold data for training a supervised critic model that mimics human ratings of the validity of contextualizations.
\section{\dataset: Dataset Design}
\label{sec:data}

\begin{figure}[t!]
    \includegraphics[width=\linewidth]{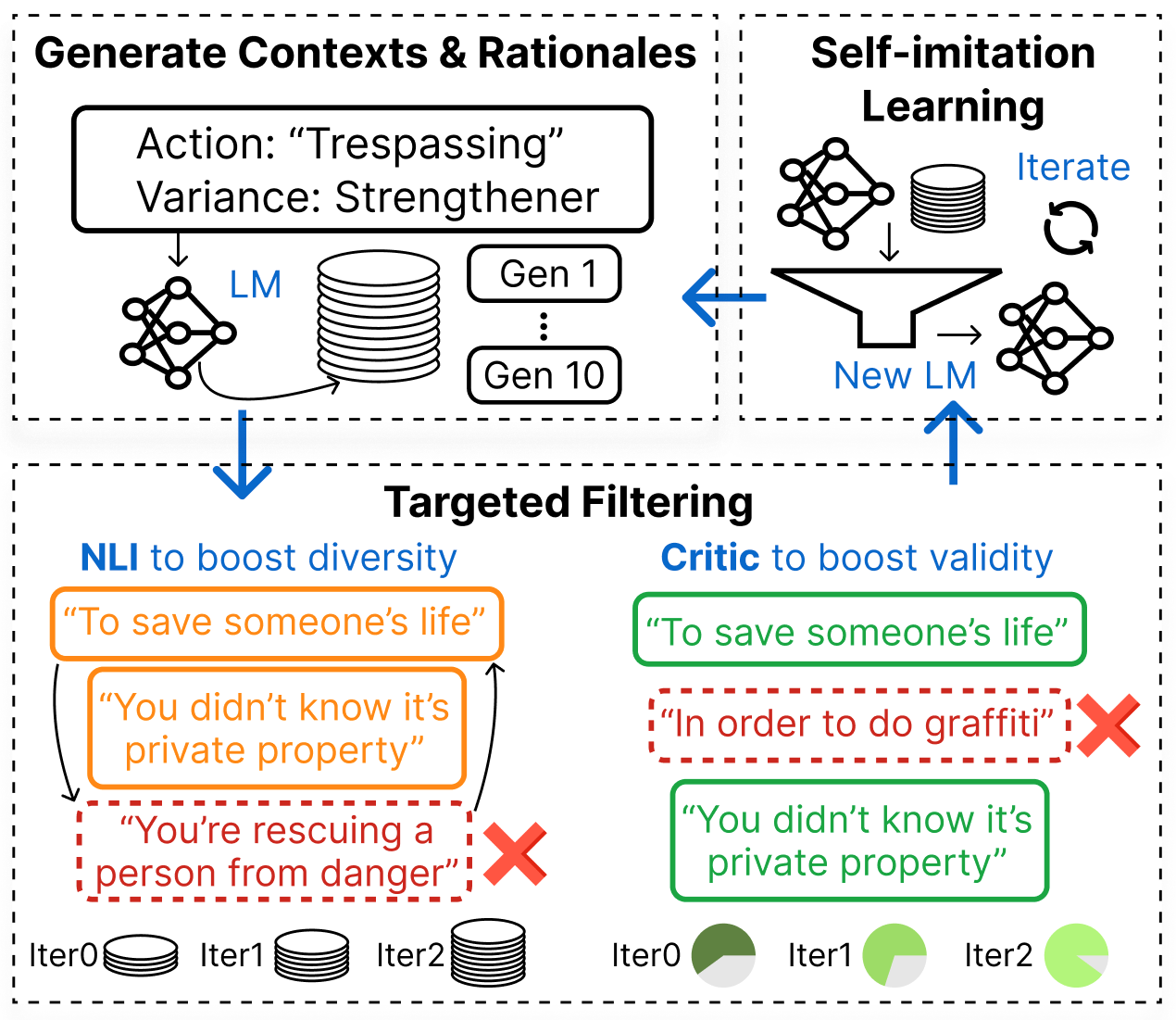}
    \caption{Iterative self-distillation that repeats generation, filtering, and self-imitation learning.}
    \label{fig:pipeline}
    \vspace{-5mm}
\end{figure}

 % \nouha{I recommend this title for the section: \dataset: Dataset Design}
 
% In this section, 
We introduce \dataset (\datasetshort), a dataset for the \textit{defeasible moral reasoning} task. 
% \datasetshort provides grounded contexts making an action more or less morally acceptable, along with commonsense rationales that justify the reasoning (see example data in Figure \ref{fig:open-fig}). 
Given an everyday action with a default commonsense moral judgment, \datasetshort captures nuances of the defeasible moral action through contextualizations that either strengthen or attenuate the acceptability of an action. It also contains rationales that explain why the contextualizations affect the judgment. See example data in Figure \ref{fig:open-fig}.

As shown in Table \ref{tab:data-stats}, \datasetshort contains 115K actions and 578K entries each of contextualizations and rationales. 
Extensive human evaluation confirms that \dataset is of high quality, demonstrating high \textit{validity} (85.9\% for contexts; 98.5\% for rationales) and \textit{language quality} reflected by fluency and grammar correctness (99.8\% for contexts; 99.7\% for rationales), on par with human written datasets \cite{west-etal-2022-symbolic}.

% Detailed statistics and human validation results of \dataset are shown in in Table \ref{tab:data-stats}).

\paragraph{Action and Commonsense Moral Judgment.}

We source our scenarios from \socialchem \cite{forbes-etal-2020-social}, a bank of rules-of-thumb (RoTs) that describe various social, cultural, and moral norms.
Each RoT consists of an action and a sociomoral judgment annotated by workers based on natural language snippets of real-life situations. 

% \textbf{Example RoT} (\underline{judgment}, \uuline{action}):\\
% ``\underline{It's dangerous} to \uuline{set a fire}''
% % from multiple online sources.

% Same example used in Social-Chem-101 paper
\begin{center}
    \textbf{Example RoT} (\underline{judgment}, \uuline{action}):\\
    ``\underline{It's dangerous} to \uuline{set a fire}''
\end{center}

% Each RoT distills a potentially complex and multi-layered situation into a single action and a sociomoral judgment on that action.
% we find them to be
% \faeze{$\leftarrow$ this can be omitted if needed space}. 
 % extend the judgments

Because RoTs combine everyday atomic actions (\eg ``set a fire'') with commonsense sociomoral judgments (\eg ``it's dangerous''), they serve as ideal seeds to be expanded with contextual nuances.

\paragraph{Morally Variant Contextualization.}

% We define \textbf{moral variance} to be a binary label, such that a contextualization with a \textit{strengthening} moral variance makes the original action more morally acceptable, and a contextualization with a \textit{weakening} moral variance has the opposite effect.
% Note that different types of actions may have different types of \textit{most relevant} morally variant contexualizations.

\textit{Moral variance} is a binary label such that \textit{strengthening} contextualizations further ground the original action to be more morally acceptable, while \textit{weakening} contextualizations have the opposite effect. Note that meaningful morally variant contextualizations range from simple properties such as locations (\eg ``in a field of dry grass'') or auxiliary actions (\eg ``when you're camping'') to complex compositional contexts with an intricate interplay between multiple atomic variations such as ``\underline{when you're camping} \uwave{in a field of dry grass}.'' Thus, we focus on eliciting flexible contexts that exercise concrete and natural effects \textit{tailored} to given actions, instead of pre-defining the categories of the contextualizations regardless of situational nuances \cite{ziems2023normbank}.

\paragraph{Commonsense Rationales.}

A critical missing piece from previous works on grounded moral judgments is rationale that ties together the actions and contextualizations by explaining the reasoning behind the defeasible effect (\eg the context ``in a field of dry grass'' might make the action ``setting a fire'' less morally acceptable ``because it's likely to burn out of control''). \dataset provides a complete picture of \textit{how} each context achieves a moral variance, paving the way toward 
% explainable and more trustworthy computational approaches to 
dissecting and understanding the varying shades of moral judgments.

\paragraph{Human Critic Gold Data.}
\label{sec:critic-gold-data}
In addition to \datasetshort, we also release a dataset of human-annotated quality assessments of machine-generated contextualizations and rationales used to train the critic model for distillation filtering (\S\ref{sec:seed-filter}).
Actions are sampled from \socialchem, and we use GPT-3 to generate contextualizations and rationales for both moral variances, each annotated by three crowdworkers. Labels are obtained by majority votes across 
annotators, but we keep only the subset with full agreement for validation and test sets to ensure high confidence in their labels. The critic gold data contains 11K actions and 20K contextualizations with quality labels across all splits.

\begin{table}[t!]
\small
\centering
    \begin{tabular}{@{}l@{\hspace{0.7\tabcolsep}}|l@{\hspace{0.7\tabcolsep}} |c@{\hspace{0.7\tabcolsep}}c@{\hspace{0.7\tabcolsep}}|r@{\hspace{0.7\tabcolsep}}r@{\hspace{0.7\tabcolsep}}}
        \toprule 

        & & \multicolumn{2}{c|}{\textbf{Statistics}} & \multicolumn{2}{c}{\textbf{Human Val.}} \\
        \midrule
            
        \textbf{Type} & \textbf{Pol.} & \textbf{\#Entry} & \textbf{\#3-Grams} & \textbf{\%Vld.} & \textbf{\%Lan.} \\
        \midrule

        % Events & - & 116322 & 110224 & - & - \\
        Action & - & 115K & 110K & - & - \\
        \midrule

        % \multirow{3}{*}{\makecell[tl]{Contexts}} & All & 577988 & 182093 & 85.9\% & 99.8\% \\
        % & More & 265894 & 107647 & 84.2\% & 100\% \\
        % & Less & 312094 & 129729 & 87.6\% & 99.6\% \\

        % \multirow{3}{*}{\makecell[tl]{Context}} & All & 578K & 182K & 85.9\% & 99.8\% \\
        % & More & 266K & 108K & 84.2\% & 100\% \\
        % & Less & 312K & 130K & 87.6\% & 99.6\% \\

        \multirow{3}{*}{\makecell[tl]{Context}} & All & 578K & 182K & 85.9 & 99.8 \\
        & Stren. & 266K & 108K & 84.2 & 100 \\
        & Weak. & 312K & 130K & 87.6 & 99.7 \\

        \midrule
        
        % \multirow{3}{*}{\makecell[tl]{Rationales}} & All & 577988 & 275089 & 98.5\%  & 99.8\% \\
        % & More & 265894 & 169796 & 98.6\% & 100\%\\
        % & Less & 312094 & 182269 & 98.4\% & 99.6\% \\

        % \multirow{3}{*}{\makecell[tl]{Rationale}} & All & 578K & 275K & 98.5\%  & 99.8\% \\
        % & More & 266K & 170K & 98.6\% & 100\%\\
        % & Less & 312K & 182K & 98.4\% & 99.6\% \\

        \multirow{3}{*}{\makecell[tl]{Rationale}} & All & 578K & 275K & 98.5 & 99.7 \\
        & Stren. & 266K & 170K & 98.6 & 99.6 \\
        & Weak. & 312K & 182K & 98.4 & 99.8 \\

        % Action & Not wanting to be friends with my ex \\
        % \midrule
        % Judgment & It's ok \\

        % \midrule

        % \multirow{2}{*}{\makecell[tl]{Positive \\ Context}} & \multirow{2}{*}{\makecell[tl]{My ex and I had a really bad breakup and \\ they are now dating someone new}}  \\

        % \\
        % \midrule
        
        % \multirow{3}{*}{\makecell[tl]{Positive \\ Rationale}} & \multirow{3}{*}{\makecell[tl]{It would be really awkward and uncomfortable \\ to be friends with my ex, especially since they \\ are dating someone new}} \\
        % \\
        % \\
        % \midrule

        % \multirow{2}{*}{\makecell[tl]{Negative \\ Context}} & \multirow{2}{*}{\makecell[tl]{My ex and I have kids together}} \\
        % \\
        % \midrule
        
        % \multirow{2}{*}{\makecell[tl]{Negative \\ Rationale}} & \multirow{2}{*}{\makecell[tl]{If we're not friends, it'll be harder to co-parent \\ our kids and it'll be confusing for them}} \\
        % \\
        
        \bottomrule
    \end{tabular}
    \caption{Statistics and human validation results of \dataset. \textbf{\#Entry} is the total number of data entries, and \textbf{\#3-Grams} is the number of unique 3-Grams of each data type. \textbf{\%Vld.} is the percentage of valid data rated by humans, and \textbf{\%Lan.} is the percentage with proper language form (\ie fluency and grammar).}
\label{tab:data-stats}
\vspace{-5mm}
\end{table}

\section{Dataset Creation via Iterative Self-distillation}
% \section{Iterative Self-distillation: Dataset Creation}
\label{sec:iterative}

% \rowcolor[gray]{0.90} 

\begin{table*}[t]
\small
\begin{tabular}{l@{\hspace{1\tabcolsep}}|r@{\hspace{1\tabcolsep}}|c@{\hspace{1\tabcolsep}}c@{\hspace{1\tabcolsep}}|c@{\hspace{1\tabcolsep}}c@{\hspace{1\tabcolsep}}c@{\hspace{1\tabcolsep}}c@{\hspace{1\tabcolsep}}|c@{\hspace{1\tabcolsep}}c@{\hspace{1\tabcolsep}}|c@{\hspace{1\tabcolsep}}c@{\hspace{1\tabcolsep}}}

            \toprule

            & & \multicolumn{6}{c|}{\textbf{Top 1 Greedy}} & \multicolumn{4}{c}{\textbf{Top 10 Sampling}} \\
            
            \midrule

            & & \multicolumn{2}{c|}{\textbf{Auto (Critic)}} & \multicolumn{4}{c|}{\textbf{Human}} & \multicolumn{2}{c|}{\textbf{Auto (Critic)}} & \multicolumn{2}{c}{\textbf{Human}} \\
            
            Model & \#Trn. & Vld. & Avg. & Vld. & Defease. & Lan. & Rationale. & \#Vld. & \#Unq. Vld. & \#Vld. & \#Unq. Vld. \\
            
            \midrule

            GPT-3 & - & 0.53 & 0.69  & 0.56 & 0.37 & 0.98 & 0.93 & - & - & - & - \\
            \bottomrule
            
            \rowcolor{cyan_c} \studentbase & 85K & 0.68 & 0.75 & 0.54 & 0.42 & 0.98 & 0.91 & 6.40 & 5.63 & 5.36 & 4.78 \\
            - No Critic & 143K & 0.60 & 0.68 & 0.51 & 0.39 & 0.98 & 0.94 & 5.63 & 5.00 & 4.66 & 4.23 \\
            \bottomrule
            
            \rowcolor{cyan_c}  \studentone & 434K & 0.75 & 0.80 & 0.60 & 0.48 & 0.97 & 0.93 &  7.08 & 5.83 & 6.04 & 5.05 \\
            - Top 1 Only & 53K & 0.71 & 0.77 & 0.59 & 0.48 & 0.98 & 0.93 & 7.06 & 3.16 & 5.74 & 2.54 \\
            - No NLI & 492K & 0.75 & 0.80 & \underline{0.64} & 0.50 & 0.98 & 0.92 & 7.12 & 5.89 & 5.93 & 4.97 \\
            \bottomrule
            
            \rowcolor{cyan_c} \studenttwo & 466K & 0.79 & \underline{0.83} & 0.62 & 0.50 & 0.98 & 0.93 & 7.60 & \underline{6.15} & \underline{6.26} & \underline{5.21} \\ 
            - Top 1 Only & 57K & 0.73 & 0.78 & 0.62 & 0.49 & 0.97 & 0.93 & 7.28 & 2.60 & 6.13 & 2.16 \\
            - No NLI & 567K & \underline{0.80} & \underline{0.83} & 0.63 & \underline{0.51} & 0.98 & 0.91 & \underline{7.65} & 5.92 & 5.93 & 4.73 \\
            - No Self-distill & 869K & 0.75 & 0.80 & 0.62 & 0.50 & 0.98 & 0.94 & 7.19 & 5.95 & 6.01 & 5.08 \\
            \bottomrule
            
            \rowcolor{cyan_c} \studentfinal & 578K & \textbf{0.86} & \textbf{0.88} & \textbf{0.71} & \textbf{0.56} & 0.99 & 0.92 & \textbf{8.40} & \textbf{6.45} & \textbf{7.26} & \textbf{5.69} \\ 

            \bottomrule
            
\end{tabular}
\caption{Automatic and human evaluation of distilled models across three iterations. We evaluate both the top 1 model generation by greedy decoding and the top 10 candidates by nucleus sampling. Best results are \textbf{bolded} and second best results are \underline{underlined} (to declutter the table, we remove the styles for Lan. and Rationale. as their results are approximately the same across all models). }
\label{tbl:main-results}
\vspace{-6mm}
\end{table*}

% \input{tables/distill_algo}

% The emergence of large language models has opened up new opportunities for automatic dataset creation through symbolic knowledge distillation

%Automatic dataset generation using LLMs (\ie symbolic knowledge distillation) has become a popular dataset creation paradigm with the increasing power of language models \cite{}. 

% The emergence of l
Competent large language models has opened up new opportunities for automatic dataset creation through symbolic knowledge distillation \cite{west-etal-2022-symbolic}. In this framework, knowledge is generated from a large teacher model, filtered to improve data quality, and then instilled into a smaller student model.
Previous works have found that machine-generated datasets can surpass human-authored datasets in their quality and diversity while also achieving greater scale
% with targeted filtering, automatically generated knowledge data 
% the quality of 
% in measures such as 
% on their quality and diversity, while also achieving greater scale 
\cite{west-etal-2022-symbolic, bhagavatula2023i2d2, sclar-etal-2022-referee, jung2023impossible, wang2023scott}.

In this work, we create \dataset with an iterative self-distillation approach which minimizes the resource bottleneck from expensive GPT-3 API calls.
Our approach follows three stages after producing an initial student model using relatively small-scale seed knowledge from GPT-3: (1) \textbf{self-distillation from student models} to move away from reliance on the expensive GPT-3 model; (2) \textbf{targeted filtering} to critically select high-quality and diverse data; and (3) \textbf{self-imitation learning} to amplify learning signals.
% from student models. 

% with a critic model that mimics human judgment
% Next, we describe each stage in detail.

% \subsection{Automatic Data Generation Pipeline}

\subsection{Gathering Medium-quality Seed Data to Train an Initial Student Model}
% \subsection{Gathering Medium-quality Seed Knowledge to Train an Initial Student Model}
\label{ssec:seed-data-gen}

\paragraph{Eliciting Raw Seed Data from GPT-3.}

% \faeze{maybe explain the high-level concept of SKD before talking about `teacher model`.}
% \kavel{added some background above}
% To set the stage for the later knowledge amplification stage process via iterative self-distillation, we gather seed knowledge from GPT-3 (175B)\footnote{\textit{text-davinci-003} is used wherever GPT-3 is mentioned} with carefully engineered task-directed prompts\footnote{See Appendix \ref{asec:gpt3-prompt} for GPT-3 prompt details.} to instill into a smaller base student model, \ie Flan-T5 (3B).

To set the stage for the later knowledge amplification process via iterative self-distillation, we gather seed knowledge from GPT-3 (175B)\footnote{\textit{text-davinci-003} is used wherever GPT-3 is mentioned}, the teacher model, to instill into Flan-T5 (3B), a smaller base student model.
To do so, we jointly generate defeasible contextualizations for moral actions and associated rationales with carefully engineered task-directed prompts.\footnote{See Appendix \ref{asec:gpt3-prompt} for GPT-3 prompt details.}
To encourage diverse alternatives of this initial seed, we generate two contexts/rationales each for both strengthening and weakening moral variances; in total, we obtain 212K contextualizations and rationales for 60K base actions.

\paragraph{Filtering Raw Seed Data with Critic Model.}
\label{sec:seed-filter}
% \paragraph{Filtering Seed Knowledge with Critic Model.}

Despite careful prompting following the task formulation, raw generations from GPT-3 remain noisy. Thus, we train a binary classifier to simulate human quality assessments on GPT-3 generated contextualizations as inspired by \citet{west-etal-2022-symbolic}.\footnote{Preliminary human evaluation results show that whenever contextualization is deemed high quality, the rationale is most likely to be high quality too (over 90\% of the time). Therefore, although some improvements could be gained on the rationales, in this work, we focus on improving the quality of contexts which starts at $\sim$50\% valid, as there's much more room to improve their quality.}
To train the critic model, we use the human quality-assessment gold labels introduced in \S\ref{sec:critic-gold-data}.
We fine-tune DeBERTa-V3 \cite{he2021deberta} on these annotations, resulting in a critic model that achieves high accuracy on a held-out validation set.\footnote{See critic model training details in Appendix \ref{assec:critic-training}}
Using the trained critic model, we filter the teacher model generations to remove errors from the distillation data and obtain an initial medium-quality training corpus, $D_0$, of 85K examples.

% \citet{west-etal-2022-symbolic} show that using a binary classifier as a critic model to filter the generations from a general language model improves the quality of a downstream distilled commonsense model.
% Similarly, we train a binary classifier to simulate human quality assessments on defeasible contextualizations on GPT-3 generated contextualizations and rationales.
% \faeze{Here is the right place to say that you filtered the dataset using your critic resulting in D consisting of X entries. (as apposed to talking about it in the next section).}

% \citet{west-etal-2022-symbolic} show that using a binary classifier as a critic model to filter the generations from a general language model improves the quality of a downstream distilled commonsense model.
% Similarly, we train a binary classifier to simulate human quality assessments on defeasible contextualizations on GPT-3 generated contextualizations and rationales.
% % \faeze{Here is the right place to say that you filtered the dataset using your critic resulting in D consisting of X entries. (as apposed to talking about it in the next section).}
% To train the critic, we use the human quality-assessment gold labels introduced in Section \ref{sec:critic-gold-data}.
% We fine-tune DeBERTa-V3 \cite{he2021deberta} on these annotations, resulting in a critic that achieves high accuracy on a held-out validation set (Appendix \ref{assec:critic-training}).

% With the trained critic model, we filter the remaining teacher model generations, which results in an initial distillation training corpus of 85K examples.

\paragraph{Training Initial Student Model}
% {with Critically Filtered Seed Knowledge.}

Our goal is to train an initial student model capable of generating contextualizations and rationales for a given action and moral variance.
We fine-tune Flan-T5 (3B) \cite{chung-et-al-2022-flant5} on 
% the filtered mid-quality distillation seed
$D_0$ for 3 epochs to produce \textbf{\studentbase}, our base student model.
%, which already performs at a similar level to the teacher model. 

\subsection{Refining Intermediate Student Models via Iterative Self-distillation}

We refine the quality of the base student model by further amplifying desired generation properties through an iterative self-distillation process, utilizing no additional generations from the teacher model.
Our iterative process has some key differences from \citet{bhagavatula2023i2d2}, in that we focus on improving \textit{diversity} in addition to quality.
% when performing self-distillation.

% \faeze{this section needs a rewrite. generally it's better to start at high-level and then move to low-level details. For example, "Our goal is to train an initial student model that is capable of generating contextualization for a given action and moral variance. Starting from our filtered dataset D, we fine-tune Flan-T5-XL for 3 epochs ... " }

% \paragraph{Self-distillation via Iterative Knowledge Models.}

% \paragraph{Learning by self-distillation.}

% From the base student knowledge model, we train two further knowledge models through an iterative self-distillation process, utilizing no additional generations from the teacher model.
% Our iterative process has some key differences from \citet{bhagavatula2023i2d2}, in that we focus on improving \emph{diversity} in addition to quality when performing self-distillation.

% \subparagraph{Learning by self-distillation.}

% \faeze{if we are going to have subparagraph, how about changing paragraphs to subsections?}
% \faeze{beam or sampling?} 

\paragraph{Self-distillation from Student Models.}

First, we generate a corpus of contextualizations and rationales using \studentbase on a set of newly sampled actions from the training split of \socialchem.
Given an action, we use nucleus sampling \cite{holtzman2020curious}  ($p = 0.9$) to produce 10 contextualizations and rationales for each moral variance. 
% We then preprocess and filter \liwei{what does preprocess and filter mean here?} this set similarly to $D_0$. 

% We use nucleus sampling with $p = 0.9$ to produce the top 10 candidates via beams search for each moral variance for a given action, and we preprocess and filter this set similarly to the initial model. \faeze{Given an action, we use nucleus sampling with $p = 0.9$ to produce 10 contextualizations for each moral variance. We then preprocess and filter this set similarly to the initial model.}

% \faeze{what input data? $D_0$? corpus?}

\paragraph{Targeted Filtering.}

Next, we again perform targeted filtering on the newly self-distilled data to (1) ensure the validity of the data via the supervised critic model, similarly to the treatment of $D_0$ described in \S \ref{ssec:seed-data-gen}; (2) encourage diverse model outputs by reducing repetition among valid contextualizations using a Natural Language Inference filter (NLI) \cite{liu-etal-2022-wanli}.

NLI is an
% classical 
NLP task that determines whether a premise statement entails or implies the truth of a hypothesis statement \cite{bowman-etal-2015-large,liu-etal-2022-wanli}. 
For a given pair of contextualizations $A$ and $B$, we say the pair is mutually entailed if both $A \to B$  and $B \to A$ are entailments, indicating a high confidence of not only lexically but also semantically repetitive content.
We filter out these mutual entailments such that at most one example from each pair remains in the dataset, thereby removing redundant signals in the training data.
We use RoBERTa-Large pretrained on the WANLI dataset \cite{liu-etal-2022-wanli} to compute entailment scores between each of the generated contextualizations for a given input. Formally, the filtering process 
% for candidates of a given action and moral variance 
is defined as:

\vspace{-8mm}
\begin{align*}
    &\text{accept}_{NLI}(c_i) = \forall j \in [1, i) : \lnot \text{accept}_{NLI}(c_j) \lor \\
    &(P_{NLI}(c_i, c_j) < 0.5 \lor P_{NLI}(c_j, c_i) < 0.5)
\end{align*}
\vspace{-8mm}

where $\text{accept}_{NLI}(c)$ determines if a context $c$ is accepted into the filtered set of unique candidates, $c_k$ is the $k$'th candidate, and $P_{NLI}(c_1, c_2)$ is the predicted score that context $c_1$ entails $c_2$.
 
% We further filter $D_i$ for iterative distillation to enforce semantic uniqueness among the valid beams for each input.
% We use RoBERTa-Large pretrained on the WANLI dataset \cite{liu-etal-2022-wanli} to compute a bidirectional entailment score between each of the generated contextualizations for a given input.
% For a given pair of contextualizations $A$ and $B$, we say the pair is mutually entailed if the predicted entailment score for both $A \to B$  and $B \to A$ are above 0.5, indicating a high confidence of not only lexical but also semantically repetitive content.
% We filter out these mutual entailments such that at most one example from each pair remains in the dataset, thereby removing redundant signals in the training data.

This process results in a filtered self-generated corpus, $D_1$, of 434K examples.
We then train \studentone using $D_1$ starting from \studentbase.

% \subparagraph{Iteration.}

To improve the student model further, we repeat the self-distillation process using \studentone as the base. Using \studentone, we generate a high-quality corpus, $D_2$, automatically filtered by the supervised critic model and NLI to train a second-iteration self-distilled student model, \studenttwo.

% is iteratively fine-tuned on its own high-quality generations, automatically identified using a supervised critic model.

% \paragraph{Rationale Quality.}
% We perform most data preparation on the contextualizations, with only minor processing of the rationales. We focus on the contextualizations because through human evaluations, we find that for both GPT-3 and downstream distilled models, the corresponding rationale to a high-quality contextualization is also judged as high-quality over 90\% of the time. Therefore, although some improvements could be gained from further optimizing the rationales, in this work we emphasize improving the contextualizations, which have a much lower baseline of quality. 
% \liwei{We don't need this paragraph as a independent subsection}

% \liwei{This paragraph could be folded under a footnote.}

% \subsection{Distilling the Final Dataset with the Best Performing Student Model}

%\subsection{Training a Final Student Model with a Large-scale, High-quality Dataset Elicited from the Best Self-distilled Model}
\subsection{Training a Final Student Model with Large-scale, High-quality Data from Refined Self-distilled Model}
% \subsection{Training a Final Student Model}

Using \studenttwo, we produce \dataset by generating contextualizations and rationales for 94K actions and combining them with previous training sets. We apply the NLI filter as above and filter by a restrictive critic model threshold of $0.96$ to ensure high confidence of quality in the dataset. Human evaluation of a subset of 1000 samples shows that 85.9\% of contextualizations and 98.5\% of rationales are deemed high quality (see details of dataset stats and human validation in Table \ref{tab:data-stats}). Using this large-scale, high-quality dataset, we train a final student model, \studentfinal, outperforming all previous intermediate student models and the teacher model (\ie GPT-3).

\begin{table}[t!]
\small
\centering
\begin{tabular}{l|ll}

            % \toprule
            % & \multicolumn{2}{c}{\textbf{Top 1 Greedy}} \\
            
            % \midrule

            % & \multicolumn{2}{c}{\textbf{Auto (Critic)}} \\
            
            \textbf{Model} & Vld. & Avg. \\
            
            \midrule

            GPT-3 (Teacher) & 0.53 & 0.69 \\
            % \midrule
            Falcon-7B-Instruct & 0.39 & 0.54 \\
            % \midrule
            GPT-3.5 (ChatGPT) & 0.71 & 0.77 \\
            % \midrule
            GPT-4 & 0.77 & 0.82 \\
            \midrule
            \studentfinal & \textbf{0.86} & \textbf{0.88} \\
            % \bottomrule
            
\end{tabular}
\caption{Our \studentfinal model outperforms all other baseline models on the top 1 generation via the automatic evaluation.}
\label{tab:additional-baselines}
\end{table}

\section{Experimentation Setups}
\label{sec:expt}

\subsection{Evaluation Data}

We hold out a test set of 6K actions from the same distribution in \socialchem.
For each model we generate contextualizations and rationales over the test set using both greedy decoding (top 1 candidate) and nucleus sampling (top 10 candidates) with $p = 0.9$.

% We aim to evaluate the overall quality of a model's generations (\textbf{validity}), a model's capability to produce diverse contextualizations for a given action and moral variance (\textbf{diversity}), and also the degree of moral variance that the model's contextualizations provide (\textbf{defeasibility}).
% % To that end, for each model we generate contextualizations and rationales over the test set using both greedy decoding and beam decoding with $p = 0.9$.

\subsection{Evaluation Metrics}

We aim to evaluate the overall quality of model-generated contextualizations and rationales (\textit{validity}), a model's capability to produce diverse contextualizations for a given action and moral variance (\textit{diversity}), and also the degree of moral variance that the model's contextualizations provide (\textit{defeasibility}).
% We evaluate the quality of the generated contextualizations on their \textit{validity}, \textit{diversity}, and \textit{defeasibility}, i
Finally, we also evaluate the general \textit{language quality} of model generations reflected by fluency and grammatical correctness.

% We collect several automated metrics and two dimensions of human annotation to measure validity and diversity of generated contextualizations.
% We evaluate automated metrics over the entire test set, and sample a subset of 200 actions for human annotation.

\paragraph{Validity.}

For \underline{contextualizations}, we use the critic model for automatic evaluation. 
For greedy generations, we compute the ratio of generations in the test set that pass the critic filter threshold as defined in \S \ref{sec:iterative} (\textbf{Auto/Vld.}) and the average predicted critic score across the test set (\textbf{Auto/Avg.}).
For sampled generations, we compute the average number of contextualizations out of the top 10 candidates that pass the critic filter threshold (\textbf{Auto/\#Vld.}). We also conduct human evaluation to vet validity of contextualizations to complement conclusions drawn from automatic evaluation (\textbf{Human/Vld., \#Vld.}). We evaluate the validity of \underline{rationales} with human evaluation (\textbf{Rationale.}).

% In human evaluations, we collect annotations for the validity of both contexts and rationales, both for greedy and beam generations, in addition to the grammatical fluency of each. For beam generations, we evaluate only the contextualization validity to reduce the task complexity for annotators presented with multiple generations at once.

% We use the critic model as the main automatic metric for the validity of model generations.
% To evaluate greedy generations, we compute the average predicted critic score and the ratio of generations in the test set that pass the critic filter threshold as defined in Section \ref{sec:iterative}.
% To assess the validity of beam generations, we compute the average number of contextualizations out of the top 10 beams for a given input that passes the critic filter threshold.

% In human evaluations, we collect annotations for the validity of both contexts and rationales, both for greedy and beam generations, in addition to the grammatical fluency of each. For beam generations, we evaluate only the contextualization validity to reduce the task complexity for annotators presented with multiple generations at once.

\paragraph{Diversity.}

We use automatic evaluation to assess the diversity of generated \underline{contextualizations} as it is a generally well-defined dimension.
Similarly to the mutual entailment NLI filter in \S \ref{sec:iterative}, we compute the bidirectional entailment probabilities between each pair of contextualizations across valid candidates and report the average number of semantically unique generations (\textbf{Auto/\#Unq. Vld.}). This metric describes the model's capability to produce multiple varied contextualizations for a given input, directly indicating the diversity of model outputs.

% \kavel{Can we say "diversity of a model"? Feels a bit anthropomorphizing}.

% This metric using human validity scores is the most direct indicator of the final quality and diversity of a model.

% We report diversity metrics for the subsets of the beams, which are scored as valid by the critic model and human annotators since diversity can be trivially increased by including invalid contextualizations.

% For lexical diversity, we compute the $\{1, 2, 3\}$-gram distinctness between the beams by the ratio between unique $n$-grams and total unique tokens.
%\kavel{remove n-gram section?}
% Similarly to the mutual entailment filter in Section \ref{sec:iterative}, we compute semantic diversity using an entailment model.
% We compute the bidirectional entailment probabilities between each pair of contextualizations across valid beams and report the average number of semantically unique generations; this metric (``Unique \& Valid beams'') describes the model's capability to produce multiple varied contextualizations for a given input. This metric using human validity scores is the most direct indicator of the final quality and diversity of a model.

\paragraph{Defeasibility.}
% \kavel{I think no auto metrics for this? Or maybe update-type classifier confidence?} 
% \liwei{could we combine automatic and human eval, but introduce the metrics by topic? \ie evaluation metrics: validity (auto, human), diversity (auto, human), defeasibility (human)}
% \kavel{That's a good idea to combine these sections, would group together the relevant topics and probably save some space too}

We break down human evaluation of contextualization validity into more granular answer choices---``significantly'' or ``slightly'' shifting the moral implication of the base action.
% we split the answer choices for contextualization validity annotations into different levels of ``significant'' and ``slight'' moral variance. 
Defeasibility of the \underline{contextualization} (\textbf{Defease.}) is computed as $(\text{\#contexts}_\text{significant} * 1 + \text{\#contexts}_\text{slight} * 0.5) / \text{\#all}$, indicating the degree to which the contextualizations affect the morality of the original actions.
See Appendix \ref{asec:human-annotation} for annotation details.

% In human evaluation, we split the answer choices for contextualization validity annotations into different levels of ``Significant'' and ``Slight'' moral variance, we also gain insight into the defeasibility of the generations; \ie the degree to which the contextualizations affect the morality of the original action.
% See Appendix \ref{asec:human-annotation} for more details and annotation interface examples.

\paragraph{Language Quality.}

We evaluate the language quality (\ie fluency and grammar correctness) of generated \underline{contexulizations} and \underline{rationales} with human evaluations (\textbf{Lan.}).

\section{Results and Discussions}
\label{sec:results}

% \section{Additional Baseline Results}
% We compute further baselines for comparison using various state-of-the-art language models to contextualize the performance of \studentfinal in the broader landscape of general-purpose LLMs. We compute only automated metrics since we have already shown in \S\ref{sec:results} that the critic metrics are correlated to human evaluations. Each additional model is evaluated using 1000 sampled inputs from the test set used to produce the metrics in Table \ref{tbl:main-results}.

% \label{asec:additional-baselines}

\begin{figure}[t!]
    \centering
    \includegraphics[width=1\linewidth]{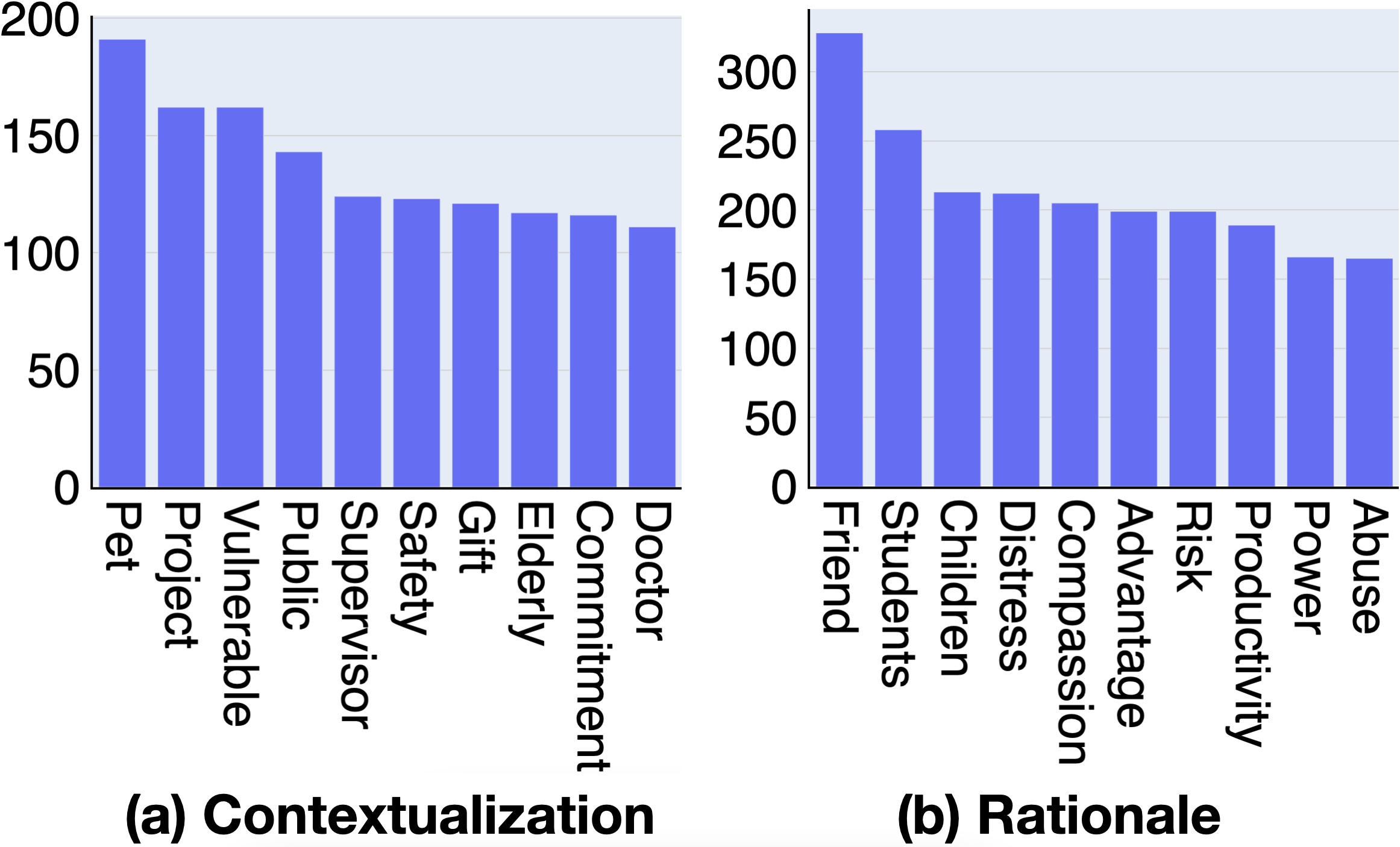}
    \caption{Top 10 topics and their counts among 10K sampled contextualizations and rationales from \datasetshort.}
    \label{fig:topics_analysis}
    \vspace{-5mm}
\end{figure}

% See Table \ref{tbl:main-results} for exact metrics of each model and ablations.
% Table \ref{tbl:main-results} shows the results 

In this section, we present results and insights of the iterative self-distillation process and an analysis of the resulting dataset, \datasetshort.

\subsection{Insights of Iterative Self-distillation}
\label{ssec:distill-results}

As shown in Table \ref{tbl:main-results}, student models improve across iterations during the iterative self-distillation process on all of \textit{validity} (0.54$\rightarrow$0.71), \textit{diversity} (4.78$\rightarrow$5.69), and \textit{defeasibility} (0.42$\rightarrow$0.56). In particular, the final student model, \studentfinal, wins over GPT-3 (the teacher model orders of magnitude larger) by a substantial relative gain on validity (26.8\%) and defeasibility (51.4\%), demonstrating the effectiveness of distilling small specialized knowledge models from large general-purpose close-sourced models like GPT-3.

% In addition to surpassing the teacher model, our distilled model also outperforms all other baseline models evaluated. Even comparing to GPT-4 (the largest available model), \studentfinal performs significantly better with almost 10\% more valid generations as determined by the critic model. We detail these supplemental baseline metrics from varied sizes of contemporary general-purpose LLMs in Appendix \ref{asec:additional-baselines}.

\paragraph{Filtering by the critic model improves the quality of contextualizations.}
Our results show that filtering training examples by the critic model improves the quality of generated contextualizations, in line with previous findings \cite{west-etal-2022-symbolic,bhagavatula2023i2d2}.
In particular, we conduct an ablation study without using critic filtering (\studentbase-\textit{No critic}), resulting in lower performance on almost all contextualization metrics and similar performance on others, despite its training set being $\sim$70\% larger than \studentbase.

% Thus, we show that the critic model filter successfully improves the data quality and, subsequentially, models trained from that data.

% . Results in Table \ref{tbl:main-results} show that \textbf{Distill$_{base}$} outperforms the ablation without critic filtering on almost 

\paragraph{Training student models on diverse self-generated data improves validity and diversity over greedy decoding.}
% We find that while self-distillation using greedy decoding improves the quality of contextualizations, it significantly decreases the diversity of the model's generations, resulting in a drop from 4.78 Unique \& Valid beams to just 2.54 in one iteration. However, when we generate training data using beam decoding to produce \textbf{SelfDistill$_{1}$}, the average number of high-quality unique beams increases in each iteration.

We find that omitting diverse candidates during the self-distillation process results in a drastic decrease in the diversity of the subsequent models. 
In particular, ablations using only the top 1 candidate in distillation (\studentone-\textit{Top 1 Only} and \studenttwo-\textit{Top 1 Only}) produce significantly less valid and unique generations compared to \studentone (5.05$\rightarrow$2.54) and \studenttwo (5.21$\rightarrow$2.16). This insight is critical as previous symbolic knowledge distillation works \cite{west-etal-2022-symbolic,bhagavatula2023i2d2} focused primarily on improving the validity of downstream student models without screening the diversity.

% We find that while self-distillation using greedy decoding improves the quality of contextualizations, it significantly decreases the diversity of the model's generations, resulting in a drop from 4.78 Unique \& Valid beams to just 2.54 in one iteration. However, when we generate training data using beam decoding to produce \textbf{SelfDistill$_{1}$}, the average number of high-quality unique beams increases in each iteration.

% \paragraph{Filtering repetitive beams by the NLI entailment score increases the diversity of the output beams of the student model.}
\paragraph{Filtering repetitions with NLI improves the diversity of candidates from student models.}

Is training on more candidates itself, without filtering out repetitions, sufficient for improving the diversity of downstream models? To answer this question, we conduct ablation studies without using the NLI mutual entailment filter, \ie \studentone-\textit{No NLI} and \studenttwo-\textit{No NLI}. Our results show that despite being trained with more data, these two models generate less valid and unique contextualizations compared to \studentone (5.05$\rightarrow$4.97) and \studenttwo (5.21$\rightarrow$4.73), shedding light on the importance of having \textit{truly} diverse training data by removing redundancy.

% In an ablation removing the NLI filter from data preparation, we find that filtering out the semantically redundant examples does improve the diversity of downstream models.
% In fact, while the first iteration without NLI filtering improves the number of unique valid beams from the base model, the second iteration in this ablation actually decreases the diversity.
% In contrast, \textbf{SelfDistill} continues to improve significantly throughout the first and second iterations.

\paragraph{Successive iterative training leads to a higher quality student model than a single iteration.}

We train an ablation model (\studenttwo-No Self-distill) combining the actions in the training sets of both the first and second rounds of distillation, but with only one iteration of self-learning from \studentbase. \studenttwo, which has been trained using the same actions over two rounds of successive training for the same number of total training steps, outperforms this ablation on almost all metrics,
% This comparison shows 
showing the effectiveness of amplifying learning signals via successive iterations of self-learning.

% We find that although the dataset size is comparable to those of \textbf{SelfDistill$_1$} and \textbf{SelfDistill$_2$} combined, \textbf{SelfDistill$_2$} outperforms this ablation on almost all metrics, including significantly improving the number of Unique \& Valid beams. This demonstrates that successive iterations of self-learning produce a higher quality model than a single round with more data.

\paragraph{The final student model outperforms orders of magnitude larger off-the-shelf models.}

We evaluate the zero-shot performance of some of the most powerful general-purpose LLMs of varied sizes with 1000 sampled examples from our test set. We use the same instructions as we used to distill the seed knowledge from GPT-3 to prompt these baselines in a zero-shot manner (see Appendix \S \ref{asec:gpt3-prompt}). Results in Table \ref{tab:additional-baselines} show that despite some of the zero-shot models being orders of magnitude larger than our final student model (\eg 175B vs. 3B), our model outperforms them all, proving the effectiveness of our proposed approach.

\subsection{Delving into \dataset}
\label{ssec:data-analysis}

\begin{figure}[t!]
    \centering
    \includegraphics[width=1\linewidth]{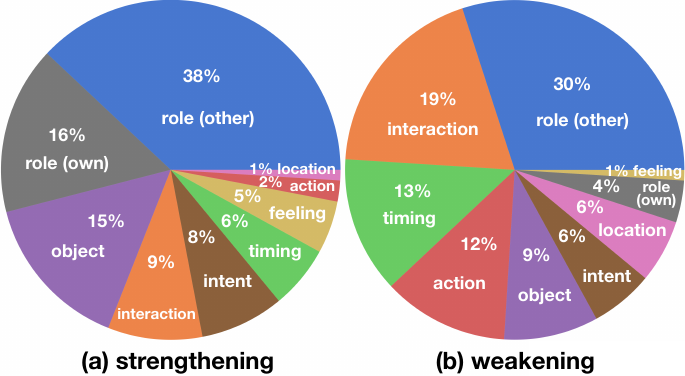}
    \caption{Qualitative analysis of contextualization categories per moral variance. (a) and (b) are for strengthening and weakening contexulizations, respectively.}
    \label{fig:more_less_ok_pie}
    \vspace{-4mm}
\end{figure}

\begin{figure}[t!]
    \centering
    \includegraphics[width=1\linewidth]{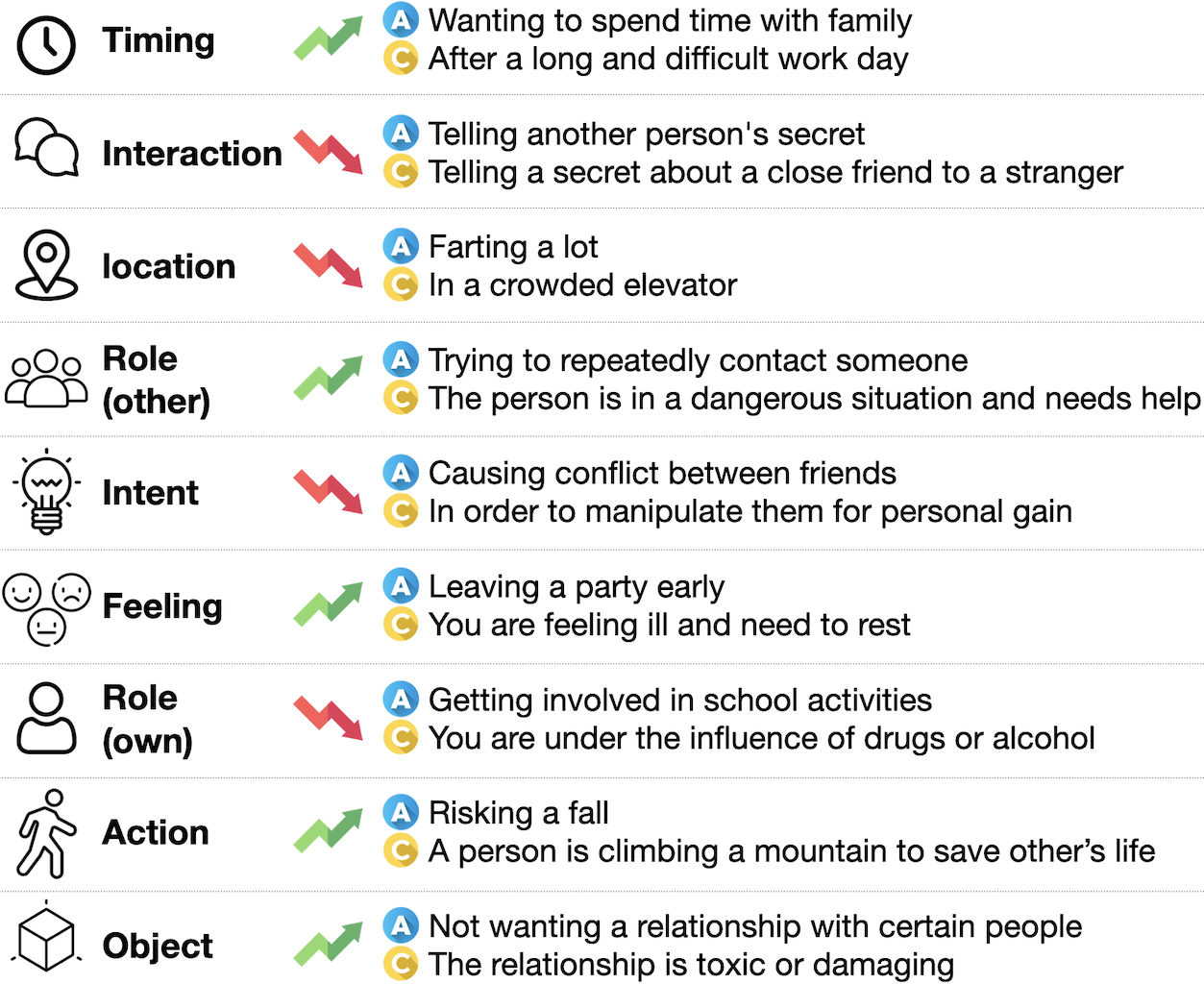}
    \caption{Example rich contextualizations per category.}
    \label{fig:more_less_ok_ex}
    \vspace{-5mm}
\end{figure}

% In this section, w
We analyze \dataset to gauge the dataset composition and gain a comprehensive picture of captured contextualizations and rationales.

% \begin{figure}[t!]
%     \begin{minipage}[t]{0.5\textwidth}
%     % \centering
%     \includegraphics[width=0.5\linewidth]{figures/more_ok_pie.png}
%     \caption{}
%     \label{fig:more_ok_pie}
%     \end{minipage}
    
%     \begin{minipage}[t]{0.48\textwidth}
%     % \centering
%     \includegraphics[width=0.48\linewidth]{figures/less_ok_pie.png}
%     \caption{}
%     \label{fig:less_ok_pie}
%     \end{minipage}
% \end{figure}

\paragraph{Contextualization.}

% \liwei{topics analysis via frequent tokens per update type}

To understand what topics contextualizations in \datasetshort 
% typically 
represent, we conduct a topic analysis with BERTopic \cite{grootendorst2022bertopic}, an easily interpretable off-the-shelf topic modeling technique. Frequent topics of contextualizations are shown in Figure \ref{fig:topics_analysis}(a), which involve \textit{daily objects or entities} (\eg pet, project, supervisor, gift, doctor) and \textit{characters or properties that carry moral weights} (\eg vulnerable, public, safety, elderly, commitment). In particular, \textit{vulnerability} serves as a critical weakening context if characters in the action are among vulnerable populations (\eg ``taking advantage of people'' becomes less acceptable if ``they are in a vulnerable state''). See topics analysis details in Appendix \S \ref{assec:topics-analysis}.

We also manually analyze the categories of 200 contextualizations sampled from \datasetshort. As shown in Figure \ref{fig:more_less_ok_pie}, frequent types of contextualizations include role specifications of the action taker (own), other characters involved in the scene (other), and setting specifications such as object, timing, and location. In addition to individual role specifications, interactions, relationships, and dynamics between multiple roles add rich groundings that carry moral significance. Figure \ref{fig:more_less_ok_ex} shows example contextualizations under each category. Contextualizations in \datasetshort have an average of 11.7 words, providing concrete, specific contexts tailored to each action.
% \liwei{Add a sentence of why interaction takes bigger part in negative contexts than positive.}
% 100 contextualizations along each moral variance direction

We further conduct a qualitative \textit{error analysis} over generations from the \studentfinal to gauge what makes a context implausible or incorrect.

\textbf{Trivial Context:} Context that adds details that may often be relevant to morality for other actions, but is rendered trivial in this particular case, \eg ``Staying up all night'' vs. ``Staying up all night while in a relationship.''

\textbf{Infeasible/Unlikely/Unnatural Context:} Context that is infeasible, highly unlikely in a real world setting, or unnatural by itself and/or in relation with the action, \eg ``Participating in a chess team'' vs. ``The team is made up of people who have been convicted of a serious crime.''

\textbf{Opposite Context:} Context that adds details opposite to the desired moral variance, \eg ``Offering people money'' vs. ``Offering money to someone who is in a vulnerable financial situation'' when prompted for a weakening context.

% \begin{itemize}
%     \item \textbf{Trivial Context:} Context adds details that may often be relevant to morality for other actions but render trivial in this particular case, \eg ``Calling-in if you're ill'' vs. ``Calling-in to work to inform your boss about your illness.''
    
%     \item \textbf{Infeasible/Unlikely/Unnatural Context:} Context is infeasible, highly unlikely in real world setting, or unnatural by itself and/or in relation with the action, \eg ``Participating in a chess team'' vs. ``The team is made up of people who have been convicted of a serious crime.''
    
%     \item \textbf{Opposite Context:} Context adds details opposite to the desired moral variance, \eg ``Offering people money'' vs. ``Offering people money to someone who is in a vulnerable financial situation'' when prompted for a weakening context.
% \end{itemize}

% \paragraph{External Factors that do:} Action is about internal feelings or emotions that are hardly defeated by external factors, \eg ``Worrying about ones you love'' vs. ``During a time of economic hardship.''
% Action is truly morally neutral (\eg internal feelings or emotions), \eg ``Enjoying hobbies'' vs. ``During a time of economic hardship.''

\paragraph{Rationales.}

% Topics analysis of rationales show that \liwei{to be continued}
% We also performed qualitative analysis into why

We conduct the same topic analysis on rationales. Results in Figure \ref{fig:topics_analysis}(b) highlight that common topics in justifying a moral decision involve \textit{important roles} (\eg friend, students, children) and \textit{common human values} (\eg distress, compassion, risk, productivity, power, abuse). See topics analysis details in Appendix \S \ref{assec:topics-analysis}.

Diving into specific examples of why contexts shift the moral implications of actions, we find common values that uplift the acceptability of an action include \textit{empathy}, \textit{kindness}, \textit{support}, and \textit{respect} (\eg ``...someone who is in need of emotional support, which shows empathy and kindness''). Additionally, \textit{(in)equality} or \textit{(un)fairness} is another dimension of value that carries significant weight on moral implications (\eg ``...taking advantage of someone else's generosity when you have the resources to provide for yourself''). Finally, contexts that are explained to promote or impede \textit{physical/mental wellbeing}, \textit{financial health}, or \textit{learning/working productivity} (\eg ``...helping them learn, which is beneficial for their future'') are also common. These qualitative results show consistency with the automatic topic analysis.

% \kavel{These sections can be condensed and streamlined. Also, should they be here or in Limitations?}

\paragraph{Toxicity Analysis.}

Since the seed actions of \datasetshort sourced from \socialchem \cite{forbes-etal-2020-social} mainly concern everyday situations, they are at a low risk of containing toxic content. In addition, due to the careful filtering with the critic model and the iteratively refined self-distillation process, we expect most of the low-quality (including potentially biased data points) to be already filtered out from the final dataset.
However, because any toxicity in moral reasoning data is especially detrimental and could easily propagate through downstream tasks, we run a toxicity analysis on a subset of 40K data points from \datasetshort using the Perspective API \cite{lees2022new}. Our results show that the average toxicity score is \textit{0.09}, indicating very low toxicity overall.
In a qualitative analysis of the data rows with higher toxicity scores (with a max of 0.83), we observe a strong pattern where the base action itself is problematic, and the distilled contexts produce the target moral variance without contributing significantly to the toxicity of the complete statement (see examples in Table \ref{tab:toxicity-examples} of Appendix \S \ref{assec:toxicity-analysis}). While no existing toxicity detection method can accurately measure all potential biases, this analysis provides reasonable confidence in the lack of toxicity in our generated contextualizations and rationales.

\paragraph{Cultural Biases}

It's also important to note the sensitivity of moral reasoning to cultural differences. In fact, previous studies have pointed out that cultural bias is a pervasive phenomenon across many NLP models (e.g., GPT-3/3.5/4) and tasks (e.g., hate speech detection with Perspective API, RewireAPI, HateRoberta) \cite{santy-etal-2023-nlpositionality}. To better represent diverse perspectives to our contextualizations, we (1) abstain from producing an absolute moral judgment given an action and (2) promote diverse distillation as discussed previously.

However, these measures cannot eliminate all traces of bias in the final model, so we also qualitatively probe \datasetshort to examine cultural biases. Admittedly, as our dataset and student models are distilled from GPT-3, which is shown to present \textit{Western-centric} perspectives, it is likely that our dataset and models inherit this cultural bias as well \cite{santurkar2023opinions, Abdulhai2023MoralFO}. 
% This Western bias is evident, f
For example, when prompted for a weakening contextualization for the action ``Not having freedom of speech in \{country\}.'' For some countries such as Japan, the United Kingdom, and the United States, the top generated context is ``in a workplace setting.'' Yet for other countries such as China, India, Thailand, Korea, and Russia, \studentfinal produces different results which might imply these countries have varying levels of human rights concerns (see details in Appendix \S \ref{assec:cultural-bias-analysis}). This example aligns with our intuition that the student model might display Western-centric biases, and it fits with a previous finding by \citet{fraser-etal-2022-moral} that such models are likely to encode the cultural biases aligned with those involved in training data annotation.

Thus, despite our careful filtering process, it is clear that culturally biased generations can still be produced by our model and may be present in \datasetshort. As such, users of this dataset must exercise discretion and care when applying it to novel use cases, and it should never be used as prescriptive ethical advice. This points to a key direction for future work to further enrich multicultural representations in computational moral reasoning and other commonsense understanding tasks.

\section{Related Work}

% \kavel{reference ProsocialDialog somewhere}
% \kavel{Related to biases and what moral ideals LLMs encode, can use RfSH missing refs 2-5}

\paragraph{Computational Morality.}
\citet{jiang2022machines} present Delphi, a commonsense moral model trained to present a descriptive view of ethical judgments. \citet{ammanabrolu2022aligning, hendrycks2021what,pmlr-v202-pan23a} incorporate moral values in an interactive game environment to align agent actions with social norms. \citet{kim-etal-2022-prosocialdialog} uses social norms to guide conversational agents' prosocial responses. \citet{jin2022make} introduce MoralExceptQA, a task of identifying the acceptability of breaking a well-known moral rule in different situations. \citet{fung2023normsage} introduce NormSAGE, a framework to discover multi-Lingual and multi-cultural norms on-the-fly. There is also a prominent line of work in quantifying social, political, and moral values and views presented in language models using well-established public opinion surveys or social science instruments \cite{santurkar2023opinions, hartmann2023political,fraser-etal-2022-moral}. Recently, \citet{sorensen2023value} builds the Kaleido model to capture the importance of pluralistic human values in moral decision-making.

\paragraph{Defeasible Reasoning.}
Defeasibility describes the idea that new information might strengthen or weaken a given interpretation \cite{rudinger-etal-2020-thinking}. This concept has been used in multiple works for different applications: \citet{rudinger-etal-2020-thinking} introduced two task formulations, one which concerns generating strengthening or weakening updates to a premise and hypothesis, and another which concerns classifying whether a premise and an update strengthen or weaken the hypothesis. \citet{madaan2021think} improved upon the latter task by modeling inference graphs. \citet{zhou2023cobra} applied defeasibility to toxic language with Contextual Bias Frames.
%\citet{allaway2022penguins} borrows from defeasible inference to reason about generics and their exceptions. % and \citet{pyatkin2022reinforced} used a defeasibility reward for generating clarification questions. 
Our work relates to recent efforts towards contextualizing moral reasoning. \citet{pyatkin2023clarifydelphi} developed ClarifyDelphi, a system capable of asking clarification questions to elicit the context surrounding a judgment.
%In contrast, in our work we directly generate contextualizations to strengthen or attenuate the morality of an action without asking specific questions.
With NormBank, \citet{ziems2023normbank} introduce a framework for grounded reasoning about %situational 
norms, adding %auxiliary information such as 
environmental conditions and agent characteristics.
Rather than %these forms of 
a QA setup or atomic groundings, %in certain categories, 
in \datasetshort, we provide free-text contextualizations along with %and we also add 
% supporting %commonsense 
rationales which justify how each piece of context alters the morality of an action.

\paragraph{Explanations and Rationales.}

% \faeze{done, see below} \liwei{ty!}

Free-form rationales have emerged as a promising direction to promote models' reasoning capabilities and aid interpretability by filling in the %commonsense reasoning or 
knowledge gap. Prior works on rationale generations take either a supervised approach by training on human-written explanations \cite{NEURIPS2018_4c7a167b,rajani-etal-2019-explain,Narang2020WT5TT,kumar-talukdar-2020-nile} or a weakly supervised approach \cite{glockner-etal-2020-think,brahman2021learning}. The advent of in-context learning (\citealp{brown2020language}; \textit{inter alia}) led to growing interest in using LLMs to generate rationales with few-shot prompting \cite{DBLP:conf/naacl/WiegreffeHSRC22,marasovic-etal-2022-shot,Wei2022ChainOT}. While such explanation-based prompting shows encouraging results, it is hindered by costly API calls. We instead endow accessible models with joint generation of contextualizations and rationales, reducing the computation required.
%To circumvent the high cost of crowdsourcing explanations, recent works relied on 
%Early works, train models on human-written rationales \cite{} or took a weakly supervised approach to genarate answers fo %, which explain model prediction in natural language, has bee

%\kavel{@Yuling} \yuling{\citet{rajani-etal-2019-explain} using dataset to train models to generate explanations -- ok Faeze added this to the paragraph already}
% \kavel{@Faeze}

\paragraph{Automatic Data Generation.}
Previous works on automatic data generation have worked on creating datasets for commonsense reasoning \cite{west-etal-2022-symbolic, bhagavatula2023i2d2, liu-etal-2022-wanli, wang2023scott, Brahman2023PlaSma}, dialogues \cite{kim2023soda, xu2023baize, koala_blogpost_2023, vicuna2023}, and summarization \cite{sclar-etal-2022-referee, jung2023impossible}.
%\citet{liu-etal-2022-wanli} introduces human-AI collaboration to produce a high-quality dataset of challenging examples.
\citet{west-etal-2022-symbolic} propose the symbolic knowledge distillation framework, and follow-up works extend it with iterative distillation \cite{sclar-etal-2022-referee, jung2023impossible, bhagavatula2023i2d2}. %and several works extend upon it with iterative distillation, \cite{sclar-etal-2022-referee, jung2023impossible, bhagavatula2023i2d2, xu2023baize}, constrained decoding \cite{bhagavatula2023i2d2}, and chain-of-thought reasoning \cite{wang2023scott}.
We further build on this paradigm to distill diversity and apply our method to 
% explainable 
moral contextualizations and rationales.

% \begin{itemize}
%     \item Commonsense: symbolic knowledge distillation \url{https://arxiv.org/abs/2110.07178}, using critic model to filter low quality data
%     \item Summarization: referee \url{https://arxiv.org/abs/2210.13800}, iterative data distillation
%     \item Summarization: Impossible Distillation \url{https://arxiv.org/abs/2305.16635}, distill from small models
%     \item Generics: i2d2 \url{https://arxiv.org/abs/2212.09246}, iterative model distillation with controllable decoding
%     \item Dialogue: SODA \url{https://arxiv.org/abs/2212.10465}, staged distillation from commonsense knowledge tuples => narratives expanded from commonsense knowledge => narrative grounded dialogues
%     \item NLI: wanli \url{https://arxiv.org/abs/2201.05955}, human-ai collaborative dataset creation
%     \item Dialogue: Baze \url{https://arxiv.org/abs/2304.01196}
%     \item Dialogue: Koala \url{https://bair.berkeley.edu/blog/2023/04/03/koala/}
%     \item Dialogue: Vicuna \url{https://lmsys.org/blog/2023-03-30-vicuna/}
%     \item SCOTT: Self-Consistent Chain-of-Thought Distillation \url{https://arxiv.org/abs/2305.01879v2}
% \end{itemize}

\section{Conclusion}

In this work, we highlight the importance of dynamic contexts in shaping moral reasoning. We introduce \textit{defeasible moral reasoning}, a task of providing grounded contexts to elucidate varying degrees of moral acceptability, accompanied by commonsense rationales to justify the reasoning. 

We employ an iterative self-distillation methodology to create a high-quality and diverse dataset, \dataset, comprising over 1.2M combined entries of contextualizations and rationales for 116K defeasible moral actions. Through this iterative approach, we also obtain a small student model capable of generating defeasible contexts with improved validity, diversity, and defeasibility.

Our work aims to promote a deeper understanding of the intricate interplay between defeasible moral actions and grounded contexts that shape moral acceptability in a nuanced and complicated way, building a strong foundation for future works on enriching cultural representations in computational moral reasoning research.
We hope \datasetshort serves as a rich resource for the community to study how moral judgments are made to unveil this unique perspective of human intelligence. 

% \kavel{(if space) deepen the discussion of future directions to lay out concrete road maps of future research on enriching cultural representations in computational moral reasoning research}

% unveiling deeper understanding how the reasoning of how moral judgments are.
% community to develop models that better simulate and assist human moral judgement in complex, real-world scenarios. 

\section*{Limitations \& Ethical Considerations}
\label{sec:ethics-limitations}

Large-language models can generate text that might be biased and insensitive to a user's socio-cultural context~\citep{bordia2019identifying,sharma2021evaluating,hovy2021five}. By introducing the \textit{defeasible moral reasoning} task, we consider different contexts and rationales, making a step towards being more diverse and inclusive in accounting for different perspectives.

However, even with our filtering by the critic model, it is possible for biased or incorrect outputs to be produced by distilled models. The critic model is trained to be a strong approximation of human judgment, but it is not perfect, and due to the scale we cannot collect human annotations to verify all examples in model training data or \dataset.

In addition, determining moral variance is a form of moral judgment and so may not have a clear answer in some cases, \eg trolley problems. There are certainly contextualizations which different groups of people disagree over whether they make the base action more or less acceptable, as could be seen in our critic gold data, where we employed inter-annotator voting to reconcile differing opinions.

With these points in mind, our dataset and models should \textit{never} be used as direct moral advice to humans; they are solely intended to be resources for more nuanced reasoning and interpretation in computational morality research.

We will publicly release our dataset and final student model to promote open-source research but for gated research purposes only. To mitigate risks of misuse, we will inform all users about potential ethical implications and risks of the resource and require them to complete a data/model user agreement form to acknowledge their consent to proper usage of the resource. This will also help us track how our resource is repurposed for downstream applications.

Finally, for extremely morally negative situations, there might not be any reasonable contextualizations to make the action justifiable (\eg ``genocide''). However, the situations we source from \socialchem focus on everyday actions that do not carry extreme moral implications. Therefore, we consider identifying impossible cases to further contextualize out of the scope of our current study. Future work could investigate more inherently morally charged cases that could not be justified even with further contextualizations.

\section*{Acknowledgement}

% This work was supported in-part by DARPA MCS program through NIWC Pacific (N66001-19-2-4031) and NSF (DMS-2134012).

The authors thank the anonymous reviewers. This research was supported in part by DARPA under the ITM program (FA8650-23-C-7316) and the Allen Institute for AI.

% \bibliography{refs}
 \bibliography{anthology,refs}
 \bibliographystyle{acl_natbib}

\clearpage
\appendix
\section{GPT-3 Prompt for Seed Data Generation}
\label{asec:gpt3-prompt}

We prompt GPT-3 with the action and moral variance to jointly generate the contextualization and rationale in a zero-shot setting. We use nucleus sampling with $p=0.9$ and presence and frequency penalties of $0.5$.

We explore multiple prompts, including few-shot versus zero-shot prompting and multiple variations of wording. From qualitative analysis, we find that this zero-shot prompt performs well, while also reducing the number of tokens required for each request.

\begin{small}
\begin{verbatim}
Given an action, write down a
situation in which the action is
more {ethical/unethical}
and give a reason for
why it makes the action
more {ethical/unethical}.
Use the following format.

Action:
Situation:
Explanation:

Action: {action}
Situation:

\end{verbatim}
\end{small}

% \section{Additional Baseline Results}
% We compute further baselines for comparison using various state-of-the-art language models to contextualize the performance of \studentfinal in the broader landscape of general-purpose LLMs. We compute only automated metrics since we have already shown in \S\ref{sec:results} that the critic metrics are correlated to human evaluations. Each additional model is evaluated using 1000 sampled inputs from the test set used to produce the metrics in Table \ref{tbl:main-results}.

% \label{asec:additional-baselines}
% \input{tables/additional_baselines}

\newpage
\newpage
\section{Human Annotation Details}
\label{asec:human-annotation}

We use Amazon Mechanical Turk as the interface for all human annotations and evaluations. For each task, we estimate the completion time by doing a selection of jobs ourselves in order to target a compensation rate of \$15 per hour.

\subsection{Critic Model Data Collection}
\label{assec:human-annotation-critic}
For the critic training data, we collect human annotations on the quality of GPT-3 generated contextualizations, which we then portion into an 80\%/10\%/10\% train/validation/test split. We combine the ``Neutral'' and ``Opposite'' answer choices into a single ``Invalid'' label.
As described in \S\ref{sec:critic-gold-data}, to reduce noise in the dataset, we collect 3 annotations per generation and vote to produce the gold label; for validation and test sets we include only cases where all three annotators agree.

\begin{figure*}
    \centering
    \includegraphics[width=0.9\linewidth]{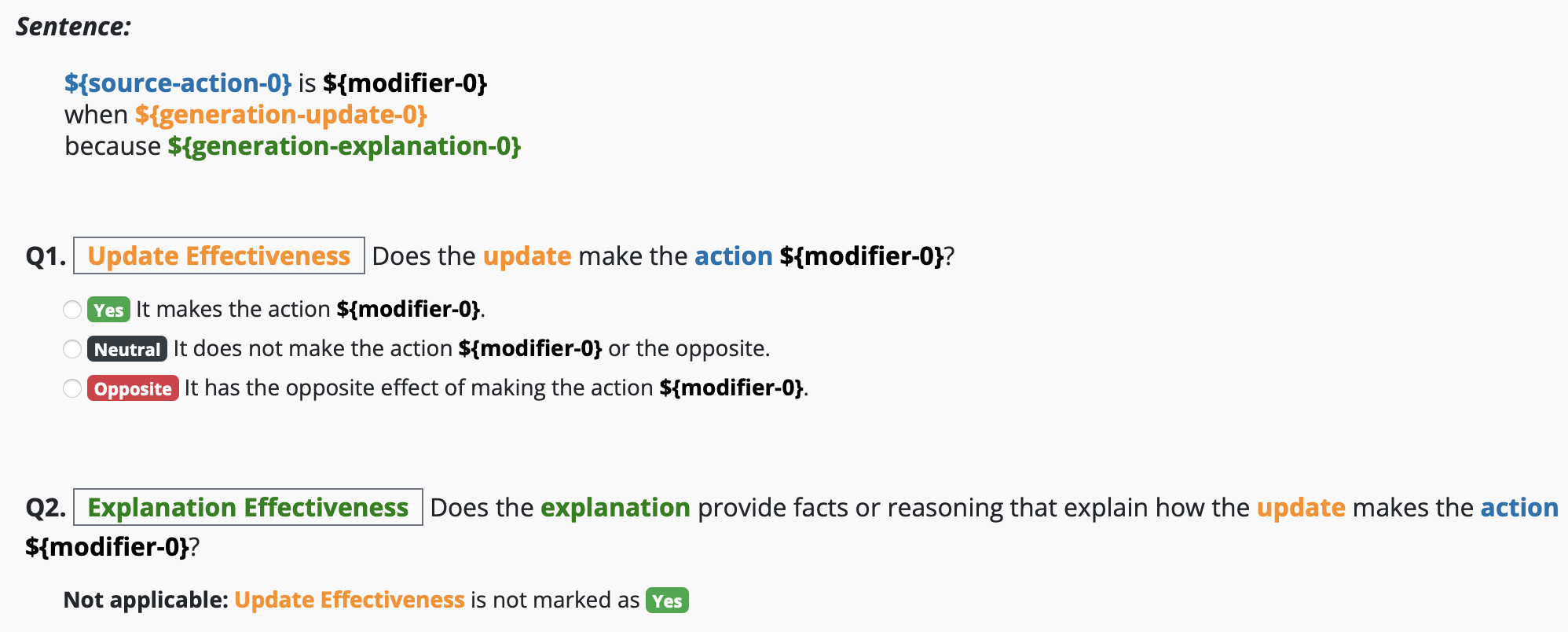}
    \includegraphics[width=0.85\linewidth]{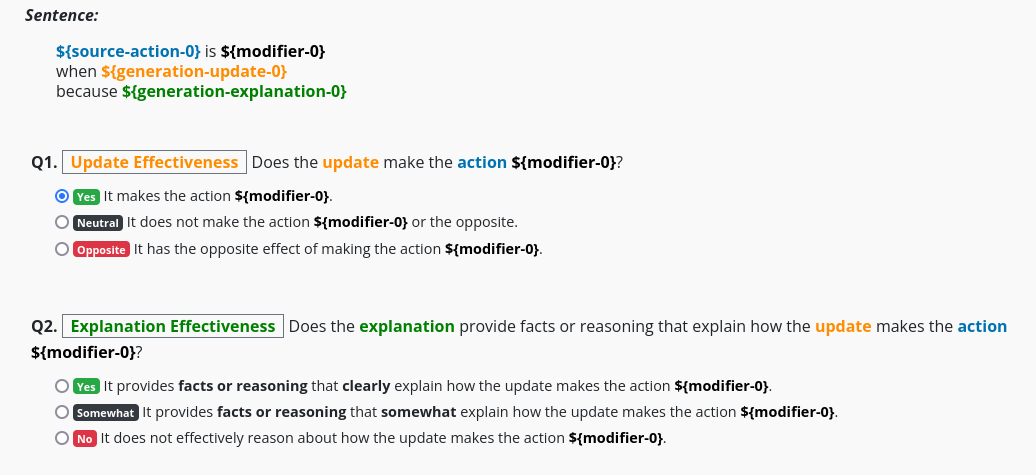}
    \caption{The human data collection template for the critic model gold training data collection.}
\end{figure*}

\subsection{Human Evaluation of Model Generations and the Final Distilled Dataset}
\label{assec:human-annotation-eval}
We design the dataset human evaluation to gather fine-grained assessments on contextualizations and rationales. As such, we include options for ``slightly'' valid contextualizations and ``somewhat'' valid rationales along with ``invalid'', which allows us to gain a more in-depth understanding of the quality of the data. In Table \ref{tab:data-stats} we collapse these labels into simply ``valid'' and ``invalid'', considering the first two options ``valid'' and only the last option ``invalid''.

\begin{table}[t!]
\small
\centering
    \begin{tabular}{@{}l@{\hspace{0.7\tabcolsep}}|l@{\hspace{0.7\tabcolsep}} |c@{\hspace{0.7\tabcolsep}}c@{\hspace{0.7\tabcolsep}}c@{\hspace{0.7\tabcolsep}}}
        \toprule 

        & & \multicolumn{3}{c}{\textbf{Human Val.}} \\
        \midrule
            
        \textbf{Type} & \textbf{Pol.} & \textbf{\%Vld.} & \textbf{\%Significant.} & \textbf{\%Slight} \\
        \midrule

        \multirow{3}{*}{\makecell[tl]{Context}} & All & 85.9 & 79.5 & 20.5 \\
        & Stren. & 84.2 & 79.0 & 21.0 \\
        & Weak. & 87.6 & 80.0 & 20.0 \\

        \midrule
        
        \multirow{3}{*}{\makecell[tl]{Rationale}} & All & 98.5 & 93.4 & 6.6 \\
        & Stren. & 98.6 & 92.9 & 7.1 \\
        & Weak. & 98.4 & 93.9 & 6.1 \\

        \bottomrule
    \end{tabular}
    \caption{Expanded human validation results of \dataset, breaking down the degree of moral variance and logical completeness. \textbf{\%Vld.} is the same as Table \ref{tab:data-stats}. \textbf{\%Significant} and \textbf{\%Slight} are (for Context) the percent of \textbf{Vld.} which significantly or slightly impact the moral variance relative to the base action; and (for Rationale) the percent of \textbf{Vld.} which are annotated as fully or somewhat valid respectively.}
\label{tab:data-stats-expanded}
\end{table}

% \textsuperscript{GOLD}

We find high inter-annotator agreement on these evaluations, with the questions on language quality and rationale validity at over 90\% full three-annotator agreement. On the context validity question, we find 57\% full agreement, and in 82\% of cases, two out of three annotators agree on a label. We expect a lower agreement on this question compared to others, since the moral judgment of the action and context is an inherently subjective task, and the answer may not be clearly defined in all cases. With the high two-way agreement, we have confidence in the accuracy of labels after applying voting across annotators.

\begin{figure*}
    \centering
    \includegraphics[width=0.9\linewidth]{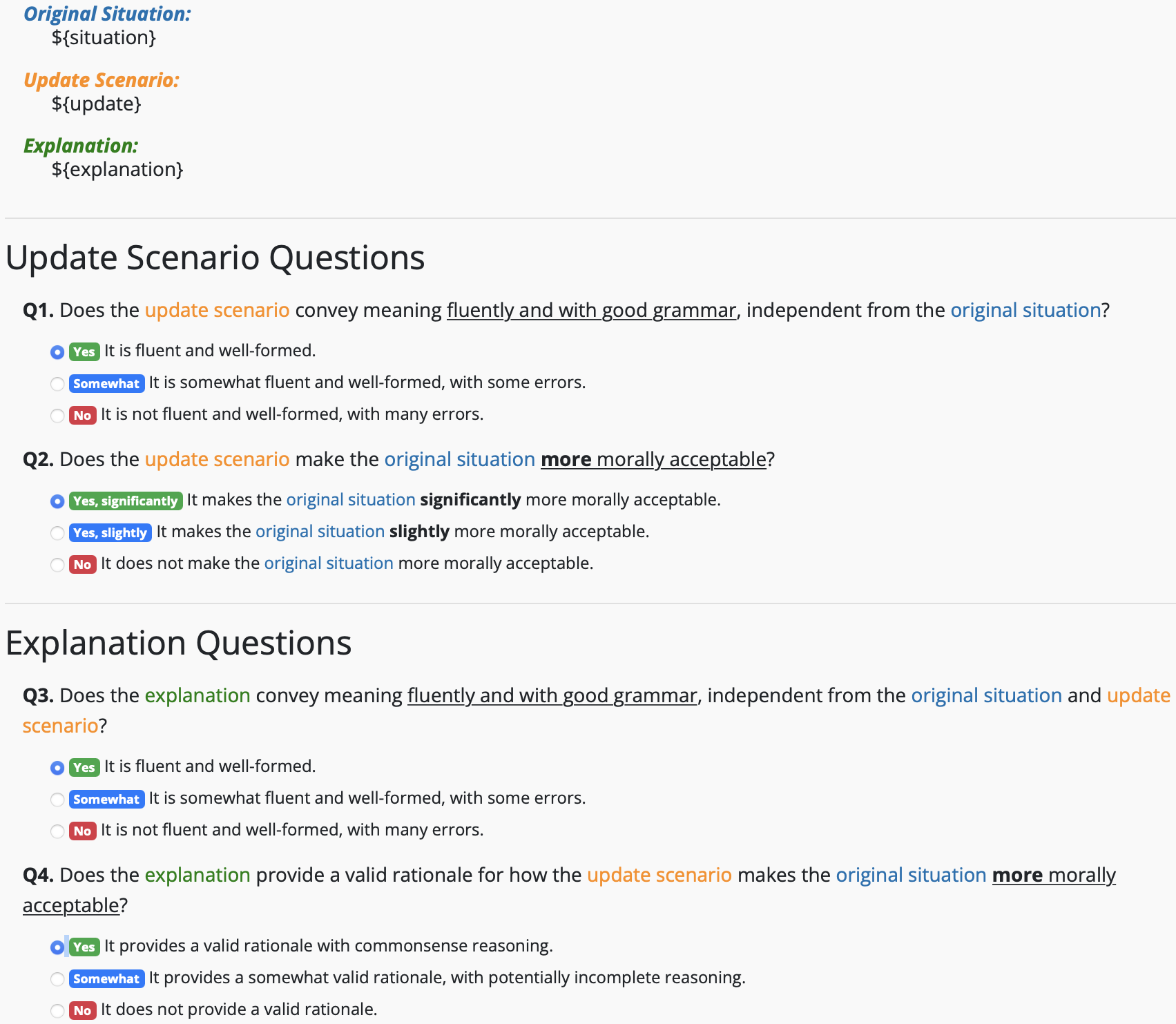}
    \caption{The human evaluation template for evaluating or validating distilled generations from teacher and student models.}
\end{figure*}

\section{Full Iterative Self-distillation Algorithm}
\label{asec:full-algo}

\begin{algorithm}[H]
\footnotesize
\label{alg:iterative-self-distillation-full}

   \begin{algorithmic}[1]

   \Require teacher model $\tau$, critic model $\rho$, $A_{\text{SocialChem}}$
   \State $A_0 \gets$ sample $A_{\text{SocialChem}}$
   \State $A_{\text{SocialChem}} \gets A_{\text{SocialChem}} \setminus A_0$
   \State $D_0 \gets$ \Call{GenerateDiverse}{$\tau$, $A_0$}
   \State $D_0 \gets$ \Call{Filter}{$\rho$, $D_0$, $\text{Threshold}_{\text{distill}}$}
   \State $\textbf{Distill}_{base} \gets$ Fine-tune base model on $D_0$
   \For {$i = 1, 2$}
       \State $A_i \gets$ sample $A_{\text{SocialChem}}$
       \State $A_{\text{SocialChem}} \gets A_{\text{SocialChem}} \setminus A_i$
       \State $\textbf{SelfDistill}_i, D_i \gets$ \Call{SelfDistill}{$\sigma$, $A_i$, $\rho$} 
   \EndFor
   \State $D_{\text{rem}} \gets$ \Call{GenerateDiverse}{$\sigma$, $A_{\text{SocialChem}}$}
   \State return \Call{Filter}{$\rho$, $D_0 \cup D_1 \cup D_2 \cup D_{\text{rem}}$, $\text{Threshold}_{\text{dataset}}$}
   \Procedure{SelfDistill}{$\sigma_{\text{old}}$, $A$, $\rho$}
        \State $D \gets$ \Call{GenerateDiverse}{$\sigma$, $A$}
        \State $D_f \gets$ \Call{Filter}{$\rho$, $D$, $\text{Threshold}_{\text{distill}}$}
        \State $\sigma_{\text{new}} \gets$ Fine-tune $\sigma_{\text{old}}$ on $D_f$
        \State return $\sigma_{\text{new}}$, $D_f$
   \EndProcedure

   \Procedure{GenerateDiverse}{$\mu$, $A$}
        \State $D \gets \emptyset$
        \For {$a$ in $A$}
            \For {$p = +, -$}
                \State $b \gets$ \{Top 10 beams from $\mu_{\text{old}}$ for $a, p$ \}
                \State $D_b \gets \emptyset$
                \For {$(a, p, c, r)$ in $b$}
                    \If {$\forall c' \in D_b: \lnot \Call{MutualNLI}{c, c'}$}
                        \State $D_b \gets D_b \cup \{ (a, p, c, r) \}$
                    \EndIf
                \EndFor
                \State $D \gets D \cup D_b$
            \EndFor
        \EndFor
        \State return $D$
    \EndProcedure
   
   \Procedure{Filter}{$\rho$, $D$, $\kappa$}
        $D_f \gets \emptyset$
        \For {$(a, p, c, r)$ in $D$}
            \If {$\rho(a, p, c) > \kappa$}
                \State $D_f \gets D_f \cup \{ (a, p, c, r) \}$
            \EndIf
        \EndFor
        \State return $D_f$
    \EndProcedure

   \Procedure{MutualNLI}{$c1$, $c2$}
        \State return $\text{Entail}(c1, c2) \land \text{Entail}(c2, c1)$
   \EndProcedure

    \end{algorithmic}
\caption{Iterative Self-distillation of \datasetshort}
\end{algorithm}

\section{Model Training Details}

\subsection{Critic Model}
\label{assec:critic-training}
We fine-tune the critic model from DeBERTa-V3-Large \cite{he2021deberta} on the human critic gold data \S\ref{sec:critic-gold-data}, attaching a 2-layer classificaton head with a hidden size of 512 and GELU \cite{DBLP:journals/corr/HendrycksG16} as the activation function.

Since the moral variance of a given contextualization may be ambiguous or contested between different annotators, we filter the validation and test data to only the subset on which all annotators agree on a label.

We conduct a small search of hyperparameters using the dev set around the values suggested by \citet{west-etal-2022-symbolic}, and we find training with batch size of 4, learning rate of $5e-06$, and dropout of 0.1 to be effective to produce a discerning critic model.
Because the dataset is heavily class-imbalanced with about 3:1 high-quality to low-quality contextualizations, we weight the loss of low-quality examples in training with the reciprocal of the imbalance.
We employ early stopping to save the checkpoint with the lowest validation loss after 15000 training steps.
We use the special tokens $\textbf{[ACTION]}$ to denote the start of the action, and $\textbf{[POS]}$ and $\textbf{[NEG]}$ respectively to denote a contextualization with positive and negative moral variance.
The critic model training takes approximately 3 hours to train on a single NVIDIA Titan XP GPU.

\begin{table}[!ht]
\centering
\begin{tabular}{ |c|c|c|c| }
    \hline
    & \textbf{Accuracy} & \textbf{F1 Score} & \textbf{AUC PR Curve} \\
    \hline
    \textbf{Val} & 0.88 & 0.93 & 0.98 \\
    \hline
    \textbf{Test} & 0.86 & 0.92 & 0.98 \\
    \hline
\end{tabular}
\caption{Critic model metrics on evaluation sets from gold human-annotated data (\S\ref{sec:critic-gold-data})}
\label{tab:critic-metrics-eval}
\end{table}

\begin{figure}[H]
    \centering
    \includegraphics[width=\linewidth]{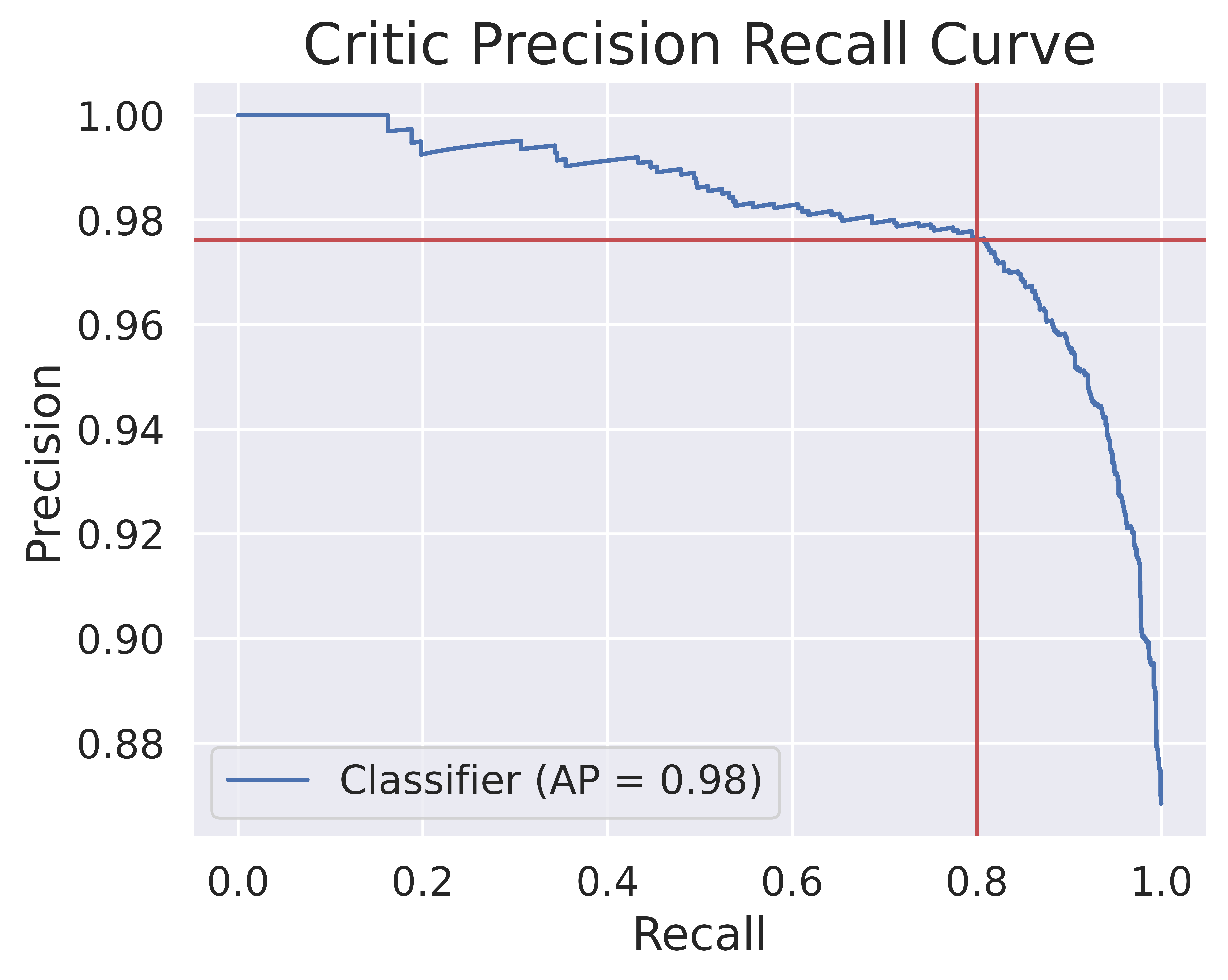}
    \caption{PR Curve on Validation Set. Red lines display recall of 0.8 with high precision used to select threshold}
    \label{fig:critic-dev-pr-curve}
\end{figure}

Using the precision-recall curve, we determine a critic threshold of $0.8$ for distillation, which achieves high precision and recall on the validation set.

\subsection{Student Model}
The base model for distillation is Flan-T5-XL \cite{chung-et-al-2022-flant5}, with 3 billion parameters.

Student model prompts follow the format:
\begin{small}
\begin{verbatim}
Action: {action}.
Modifier: more {ethical/unethical}.
\end{verbatim}
\end{small}
And target generations follow this format:
\begin{small}
\begin{verbatim}
Update: {context}. Explanation: {rationale}.

\end{verbatim}
\end{small}

We fine-tune each student model for a fixed 3 epochs using a maximum target length of 512, per-device batch size of 8, and learning rate of $5e^{-5}$. We use the AdamW optimizer, keeping other hyperparameters at default. Each student model takes about 28 hours to fine-tune using 8 NVIDIA RTX A6000 GPUs.

\section{Data Analysis}

\subsection{Dataset Examples and Topics Analysis}
\label{assec:topics-analysis}

We show examples from the \dataset in Table \ref{tab:example-data-appendix}, and common topics of contexualizations in Table \ref{tab:context-topics-appendix} and of rationales in Table \ref{tab:rationale-topics-appendix}.

\begin{table}[t]
\small
\centering
    \begin{tabular}{l @{}|l @{}}
        \toprule 
        \textbf{Attribute} & \textbf{Content} \\
        \toprule

        \rowcolor{orange} \textbf{Action} & Not wanting to be friends with my ex \\
        \midrule
        Judgment & It's ok \\

        \midrule

        \multirow{2}{*}{\makecell[tl]{Strengthen \\ Context}} & \multirow{2}{*}{\makecell[tl]{My ex and I had a really bad breakup and \\ they are now dating someone new}}  \\

        \\
        \midrule
        
        \multirow{4}{*}{\makecell[tl]{Strengthen \\ Rationale}} & \multirow{4}{*}{\makecell[tl]{It would be really awkward and \\ uncomfortable to be friends with my ex, \\ especially since \\ they are dating someone new}} \\
        \\
        \\
        \\
        \midrule

        \multirow{2}{*}{\makecell[tl]{Weaken \\ Context}} & \multirow{2}{*}{\makecell[tl]{My ex and I have kids together}} \\
        \\
        \midrule
        
        \multirow{2}{*}{\makecell[tl]{Weaken \\ Rationale}} & \multirow{2}{*}{\makecell[tl]{If we're not friends, it'll be harder to co-parent \\ our kids and it'll be confusing for them}} \\
        \\
        \midrule
        % \\
        
        % \midrule
        
        \rowcolor{orange} \textbf{Action} & Letting your mom borrow your car \\
        \midrule
        Judgment & It's nice \\

        \midrule

        \multirow{2}{*}{\makecell[tl]{Strengthen \; \\ Context}} & \multirow{2}{*}{\makecell[tl]{Your mom is unable to afford a car and \\ needs transportation for her job interview}}  \\

        \\
        \midrule
        
        \multirow{2}{*}{\makecell[tl]{Strengthen \\ Rationale}} & \multirow{2}{*}{\makecell[tl]{It demonstrates kindness and generosity \\ towards your mother}} \\
        \\
        \midrule

        \multirow{2}{*}{\makecell[tl]{Weaken \\ Context}} & \multirow{2}{*}{\makecell[tl]{Your mom is driving under the influence of \\ drugs or alcohol}} \\
        \\
        \midrule
        
        \multirow{3}{*}{\makecell[tl]{Weaken \\ Rationale}} & \multirow{2}{*}{\makecell[tl]{It increases the risk of her driving while \\ impaired, which could put other people \\ in danger}} \\
        \\
        \\
        \midrule
        % \\
        
        % \midrule
        
        \rowcolor{orange} \textbf{Action} & Flaking out on someone \\
        \midrule
        Judgment & It's rude \\

        \midrule

        \multirow{2}{*}{\makecell[tl]{Strengthen \\ Context}} & \multirow{2}{*}{\makecell[tl]{The person has been acting in a way that is \\ damaging to themselves or someone else}}  \\

        \\
        \midrule
        
        \multirow{2}{*}{\makecell[tl]{Strengthen \\ Rationale}} & \multirow{2}{*}{\makecell[tl]{It is an act of self-preservation and ensuring \\ that the person's safety is prioritized}} \\
        \\
        \midrule

        \multirow{3}{*}{\makecell[tl]{Weaken \\ Context}} & \multirow{2}{*}{\makecell[tl]{Flaking out a close friend who is going \\ through a difficult time and you know that \\ they need your support}} \\
        \\
        \\
        \midrule
        
        \multirow{2}{*}{\makecell[tl]{Weaken \\ Rationale}} & \multirow{2}{*}{\makecell[tl]{It can be seen as taking advantage of their \\ vulnerability and disregarding their feelings}} \\
        \\
        \midrule
        % \\
        
        % \midrule
        
        \rowcolor{orange} \textbf{Action} & Buying lottery tickets at the store \\
        \midrule
        Judgment & It's common \\

        \midrule

        \multirow{2}{*}{\makecell[tl]{Strengthen \\ Context}} & \multirow{2}{*}{\makecell[tl]{The lottery tickets are bought in order to \\ support a charitable cause}}  \\

        \\
        \midrule
        
        \multirow{3}{*}{\makecell[tl]{Strengthen \\ Rationale}} & \multirow{3}{*}{\makecell[tl]{It supports a good cause and helps to \\ raise money for a cause that can benefit \\ those in need}} \\
        \\
        \\
        \midrule

        \multirow{2}{*}{\makecell[tl]{Weaken \\ Context}} & \multirow{2}{*}{\makecell[tl]{Buying lottery tickets at the store to cover \\ up a theft}} \\
        \\
        \midrule
        
        \multirow{2}{*}{\makecell[tl]{Weaken \\ Rationale}} & \multirow{2}{*}{\makecell[tl]{It is enabling illegal activity}} \\
        % \\
        % \midrule
        % \\
        
        % \midrule
        
        % \rowcolor{maroon} \textbf{Action} & Taking pride in your accomplishments \\
        % \midrule
        % Judgment & It's ok \\

        % \midrule

        % \multirow{2}{*}{\makecell[tl]{Strengthen \\ Context}} & \multirow{2}{*}{\makecell[tl]{You have worked hard for a long period of time \\ to complete a project}}  \\

        % \\
        % \midrule
        
        % \multirow{3}{*}{\makecell[tl]{Strengthen \\ Rationale}} & \multirow{3}{*}{\makecell[tl]{It rewards the effort that you put in and \\ encourages you to continue working hard}} \\
        % \\
        % \\
        % \midrule

        % \multirow{2}{*}{\makecell[tl]{Weaken \\ Context}} & \multirow{2}{*}{\makecell[tl]{Your accomplishments were achieved by exploiting \\ the labor of vulnerable people}} \\
        % \\
        % \midrule
        
        % \multirow{2}{*}{\makecell[tl]{Weaken \\ Rationale}} & \multirow{2}{*}{\makecell[tl]{It perpetuates an unequal power dynamic, and \\ reinforces a cycle of exploitation}} \\
        \\
        
        \bottomrule
    \end{tabular}
\caption{Example data from \datasetshort.}
\label{tab:example-data-appendix}
\end{table}

\begin{table}[t]
\small
\centering
    \begin{tabular}{l @{}|l @{}}
        \toprule 
        \textbf{Count } & \textbf{Topic} \\
        \midrule
        191 & Pet \\
        \midrule
        162 & project \\
        \midrule
        121 & gift \\
        \midrule
        124 & supervisor \\
        \midrule
        143 & public \\
        \midrule
        111 & doctor \\
        \midrule
        90 & siblings \\
        \midrule
        93 & race \\
        \midrule
        94 & food \\
        \midrule
        162 & vulnerable \\
        \midrule
        123 & safety \\
        \midrule
        109 & decision \\
        \midrule
        84 & teacher \\
        \midrule
        95 & minor \\
        \midrule
        90 & ex \\
        \midrule
        116 & committed \\
        \midrule
        86 & friends \\
        \midrule
        98 & years \\
        \midrule
        86 & stealing \\
        \midrule
        117 & elderly \\
        \midrule
        93 & abusive \\
        \midrule
        99 & relationship \\
        \midrule
        80 & pandemic \\
        \midrule
        85 & overwhelmed \\
        \midrule
        59 & roommate \\
        \midrule
        98 & mental health \\
        \midrule
        88 & power \\
        \midrule
        98 & child \\
        \midrule
        53 & interview \\
        \midrule
        78 & neglecting \\
        \midrule
        93 & harm \\
        \midrule
        59 & expensive \\
        \midrule
        50 & gatherings \\
        \midrule
        55 & married \\
        \midrule
        72 & stranger \\
        \midrule
        65 & workplace \\
        \midrule
        79 & protect \\
        \midrule
        71 & advice \\
        \midrule
        64 & parents \\
        \midrule
        51 & formal \\
        \midrule
        60 & consistently \\
        \bottomrule
    \end{tabular}
    \caption{Topics and their Appearance Counts for 10K Sampled Contextualizations from \datasetshort, supplementing Figure \ref{fig:topics_analysis}}
    \label{tab:context-topics-appendix}
\end{table}
\begin{table}[t]
\small
\centering
    \begin{tabular}{l @{}|l @{}}
        \toprule 
        \textbf{Count } & \textbf{Topic} \\
        \midrule
        328 & friends \\
        \midrule
        258 & students \\
        \midrule
        213 & children \\
        \midrule
        199 & taking advantage \\
        \midrule
        121 & family \\
        \midrule
        144 & commitment \\
        \midrule
        124 & partner \\
        \midrule
        118 & vulnerability \\
        \midrule
        88 & marriage \\
        \midrule
        121 & care \\
        \midrule
        88 & interview \\
        \midrule
        134 & protect \\
        \midrule
        72 & pet \\
        \midrule
        87 & virus \\
        \midrule
        88 & parent \\
        \midrule
        69 & animals \\
        \midrule
        76 & discrimination \\
        \midrule
        71 & elderly \\
        \midrule
        91 & respect \\
        \midrule
        89 & helping \\
        \midrule
        157 & mental health \\
        \midrule
        88 & informed \\
        \midrule
        96 & decision \\
        \midrule
        77 & minors \\
        \midrule
        64 & endangering \\
        \midrule
        101 & financial hardship \\
        \midrule
        62 & customer \\
        \midrule
        75 & violates trust \\
        \midrule
        60 & motivated \\
        \midrule
        68 & hard work \\
        \midrule
        60 & safety \\
        \midrule
        68 & self-preservation \\
        \midrule
        80 & autonomy \\
        \midrule
        59 & relax \\
        \midrule
        55 & growth \\
        \midrule
        110 & resources \\
        \midrule
        49 & pedestrians \\
        \midrule
        57 & overspending \\
        \midrule
        105 & support \\
        \midrule
        60 & colleagues \\
        \midrule
        61 & exploitation \\
        \bottomrule
    \end{tabular}
    \caption{Common Topics and their Appearance Counts for 10K Sampled Rationales from \datasetshort, supplementing Figure \ref{fig:topics_analysis}}
    \label{tab:rationale-topics-appendix}
\end{table}

\subsection{Toxicity Analysis}
\label{assec:toxicity-analysis}

Table \ref{tab:toxicity-examples} shows examples of \dataset with high toxicity rated by the Perspective API. Essentially, all data with higher toxicity scores are cases where the base action itself is inappropriate, and the distilled contexts with the desired moral variance do not contribute significantly to the toxicity of the total statement.

\begin{table*}[t]
\small
\centering
    \begin{tabular}{l l l}
        \toprule 
        \textbf{Action} & \textbf{Moral Variance} & \textbf{Content} \\
        \toprule

        Telling someone to f*ck off & Weakener & in a professional setting \\
        % & Strengthener & Your friend has been being verbally abusive and manipulative to you for a long time \\
        \midrule

        \multirow{2}{*}{\makecell[tl]{Telling someone to f*ck off}} & \multirow{2}{*}{\makecell[tl]{Strengthener}} & \multirow{2}{*}{\makecell[tl]{Your friend has been being verbally abusive and \\ manipulative to you for a long time}} \\
        \\
        \midrule
        
        Engaging in bullshit & Strengthener & in order to protect a vulnerable person from harm \\
        \midrule
        Touching your friend's butt & Weakener & while they are in the middle of a private conversation \\

        \bottomrule
    \end{tabular}
\caption{Example data with high toxicity rated by the Perspective API.}
\label{tab:toxicity-examples}
\end{table*}

\subsection{Cultural Bias Analysis}
\label{assec:cultural-bias-analysis}

As it’s extremely challenging to systematically quantify cultural biases with state-of-the-art tools such as hate speech detectors, we thus have probed the model qualitatively to gauge evidence of cultural biases inherent in the dataset/model. Table \ref{tab:cultural-bias-examples} Shows examples generated by the final distilled student model that potentially implies cultural biases. We can see that the model indeed comes back with different updates for the prompt ``Not having freedom of speech in \{country\}'' for different countries. For some countries such as Japan, the United Kingdom, and the United States, the generated weakener context is ``in a workplace setting.'' Yet, for other countries such as China, India, Thailand, and Korea, or Russia, the model comes back with different results, which might imply these countries have varying levels of human rights concerns. This example confirms our intuition that the student model might encode Western-centric biases.

\begin{table*}[t]
\small
\centering
    \begin{tabular}{l l}
        \toprule 
        \textbf{Country} & \textbf{Context} \\
        \toprule
        
        China & In a situation where the government is using its power to oppress citizens  \\
        \midrule
        India & In a country where people are expressing their opinions on controversial topics \\
        \midrule
        Thailand & A country with a history of human rights abuses \\
        \midrule
        Korea & A country with a history of human rights abuses \\
        \midrule
        Russia & In a country with a history of human rights abuses \\
        \midrule
        Japan & In a workplace setting \\
        \midrule
        United Kingdom & In a workplace setting \\
        \midrule
        United States & In a workplace setting \\
        \bottomrule
    \end{tabular}
\caption{Examples that potentially imply cultural biases generated by the final distilled student model, for the action - ``Not having freedom of speech in \{country\}'' and the moral variance - ``weakener.''}
\label{tab:cultural-bias-examples}
\end{table*}

\end{document}